\PassOptionsToPackage{hidelinks,hyperfootnotes=false}{hyperref} 

\documentclass[12pt]{article}
\usepackage[table]{xcolor} 
\usepackage{comment}
\usepackage{tabularx}
\usepackage{arxiv}
\usepackage{longtable}
\usepackage{array}
\usepackage{makecell}
\usepackage{adjustbox}
\usepackage{enumitem}
\usepackage{colortbl} 
\usepackage{multirow}
\usepackage[utf8]{inputenc} 
\usepackage[T1]{fontenc}    

\usepackage[hyphens]{url}

\usepackage{url}            
\usepackage{booktabs}       
\usepackage{amsfonts}       
\usepackage{nicefrac}       
\usepackage{microtype}      
\usepackage{lipsum}		
\usepackage{graphicx}
\usepackage{natbib}
\usepackage{doi}
\usepackage{longtable}
\usepackage{array}
\usepackage{longtable}
\usepackage{array} 
\usepackage{geometry}
\usepackage{pifont}
\usepackage{tikz}
\usetikzlibrary{arrows.meta, positioning, fit, calc}
\usepackage{booktabs}            
\usepackage{etoolbox}
\usepackage{hyperref}

\usepackage{amssymb}                 

\usepackage{tablefootnote}
 
\definecolor{verylightgray}{gray}{0.5}

\usepackage{pifont}

\newcommand{\whitebox}{\tikz\filldraw[fill=white, draw=black] (0,0) rectangle (0.25,0.25);}
\newcommand{\blackbox}{\tikz\filldraw[fill=black, draw=black] (0,0) rectangle (0.25,0.25);}
\newcommand{\graybox}{\tikz\filldraw[fill=verylightgray, draw=verylightgray] (0,0) rectangle (0.25,0.25);}

\newif\ifshowcomments
\showcommentstrue
\ifshowcomments
  \newcommand{\KL}[1]{\textbf{\textcolor{blue}{{{KL: \#\arabic{noteXXctr}: }}#1}} \addtocounter{noteXXctr}{1}}
\else
  \renewcommand{\KL}[1]{}
\fi
\newcounter{noteXXctr} 

\title{Deep Research Agents:\\A Systematic Examination And Roadmap}
\renewcommand{\shorttitle}{Deep Research Agents: A Systematic Examination And Roadmap}

\usepackage{authblk}            
          
\usepackage[symbol]{footmisc}

\newcommand{\equalmark}{\textsuperscript{†}}  
\newcommand{\corrmark}{\textsuperscript{‡}}

\makeatletter
\renewcommand\AB@affilnote[1]{}  
\makeatother

\author[1]{Yuxuan~Huang\equalmark}
\author[2]{Yihang~Chen\equalmark}
\author[2]{Haozheng~Zhang\equalmark}
\author[3]{Kang~Li}
\author[4]{Huichi~Zhou}
\author[1]{Meng~Fang}
\author[4]{Linyi~Yang}
\author[2]{\authorcr Xiaoguang~Li}
\author[2]{Lifeng~Shang}
\author[2]{Songcen~Xu}
\author[2]{Jianye~Hao}
\author[2]{Kun~Shao\corrmark}
\author[4]{Jun~Wang\corrmark}

\affil[]{\parbox{\textwidth}{
  \centering \textsuperscript{1}\,University of Liverpool \quad \textsuperscript{2}\,Huawei Noah’s Ark Lab   \quad  \textsuperscript{3}\,University of Oxford \quad \textsuperscript{4}\,University College London}}

\date{}

\usepackage{geometry}
\begin{document}
\maketitle

\begin{abstract}

The rapid progress of Large Language Models (LLMs) has given rise to a new category of autonomous AI systems, referred to as Deep Research (DR) agents. These agents are designed to tackle complex, multi-turn informational research tasks by leveraging a combination of dynamic reasoning, adaptive long-horizon planning, multi-hop information retrieval, iterative tool use, and the generation of structured analytical reports. In this paper, we conduct a detailed analysis of the foundational technologies and architectural components that constitute Deep Research agents. We begin by reviewing information acquisition strategies, contrasting API-based retrieval methods with browser-based exploration. We then examine modular tool-use frameworks, including code execution, multimodal input processing, and the integration of Model Context Protocols (MCPs) to support extensibility and ecosystem development. To systematise existing approaches, we propose a taxonomy that differentiates between static and dynamic workflows, and we classify agent architectures based on planning strategies and agent composition, including single-agent and multi-agent configurations. We also provide a critical evaluation of current benchmarks, highlighting key limitations such as restricted access to external knowledge, sequential execution inefficiencies, and misalignment between evaluation metrics and the practical objectives of DR agents. Finally, we outline open challenges and promising directions for future research. A curated and continuously updated repository of DR agent research is available at: \href{https://github.com/ai-agents-2030/awesome-deep-research-agent} {https://github.com/ai-agents-2030/awesome-deep-research-agent}.

\end{abstract}

\section{Introduction}

Recent advances in large language models (LLMs) have led to the rapid emergence of sophisticated AI agents capable of autonomous research. Early models such as GPT-3~\citep{brown2020languagemodelsfewshotlearners} primarily addressed isolated tasks, including question answering and machine translation. Subsequently, integration with external tools enabled models such as WebGPT~\citep{nakano2021webgpt} to navigate the web and synthesise information from diverse sources autonomously. Most recently, a new class of advanced autonomous systems, termed Deep Research (DR) agents, has emerged, exemplified by industry-leading solutions such as OpenAI DR~\citep{openai2025deepresearch}, Gemini DR~\citep{geminideepresearch}, Grok DeepSearch~\citep{grokdeepresearch}, and Perplexity DR~\citep{perplexitydeepresearch}. These deep research agents significantly extend LLMs by incorporating advanced reasoning, dynamic task planning, and adaptive interaction with web resources and analytical tools.

Formally, we define \textbf{``Deep Research Agents''} as:

\begin{center}
\adjustbox{max width=\textwidth}{%
    \parbox{0.94\textwidth}{\normalsize\centering
 \textit{AI agents powered by LLMs, integrating dynamic reasoning, adaptive planning, and iterative tool use to acquire, aggregate, and analyse external information, culminating in comprehensive outputs for accomplishing open-ended informational research tasks.}%
    }
}
\end{center} 

Specifically, DR agents leverage LLMs as their cognitive core, retrieving external knowledge in real-time through web browsers and structured APIs, and dynamically invoking analytical tools via customised toolkits or standardised interfaces such as the Model Context Protocol (MCP). This architecture enables DR agents to autonomously manage complex, end-to-end research workflows by seamlessly integrating reasoning processes with multimodal resources.

\begin{figure}[!t]
    \centering
    \includegraphics[width=\linewidth]{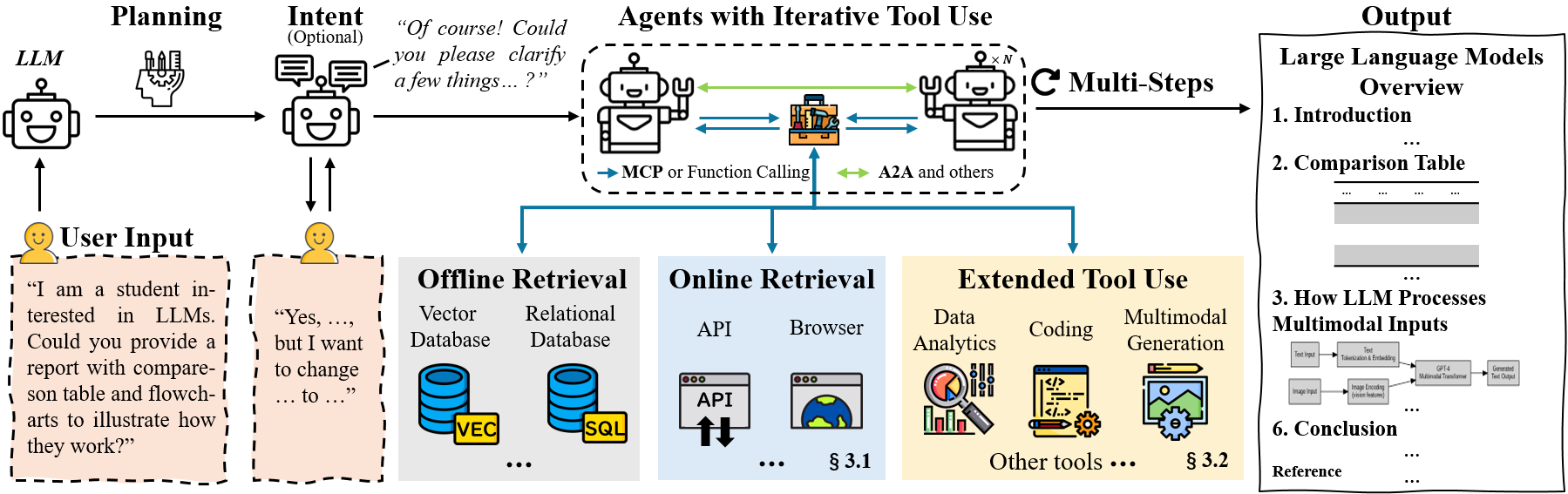}
    \caption{A structural overview of a DR agent in a multi-agent architecture for ease of illustration.}
    \label{fig1}
\end{figure}

Compared with traditional Retrieval-Augmented Generation (RAG) methods~\citep{singh2025agentic}, which primarily enhance factual accuracy but lack sustained reasoning capabilities~\citep{chen2025improving}, and conventional Tool Use (TU) systems~\citep{qu2025tool} that heavily depend on pre-defined workflows~\citep{wang2025tdag}, DR agents offer significantly greater autonomy, continual and deep reasoning abilities, dynamic task planning, and adaptive real-time interaction. These advanced capabilities uniquely position DR agents to handle complex, evolving, and knowledge-intensive research scenarios. A representative example of such a DR agent architecture is illustrated in Figure~\ref{fig1}, which demonstrates the complete workflow from user input through optional planning and intent clarification, to iterative tool utilization encompassing offline retrieval (vector and relational databases), online retrieval (APIs and browsers), and extended capabilities including data analytics, coding (etc.), and multimodal generation, ultimately producing comprehensive structured report.

\paragraph{Contribution.}
This survey systematically reviews recent advancements in DR agents, providing a comprehensive analysis of core technologies, methodologies, optimisation pipelines, and representative implementations. Specifically, the contributions of this survey include:

\begin{itemize}
\item A thorough analysis of representative DR systems, explicitly examining their system architectures, retrieval mechanisms, tool invocation methods, and performance characteristics, alongside optimisation and tuning paradigms.
\item A unified classification framework (Figure \ref{fig:workflow}) that systematically categorises DR systems based on workflow characteristics (static versus dynamic), planning strategies, and agent-based architectures (single-agent versus multi-agent), bridging diverse technical methodologies and current industrial solutions.

\item A systematic review and categorisation of existing benchmarks utilised to evaluate DR systems, highlighting how these benchmarks assess critical capabilities, such as retrieval accuracy, reasoning depth, and adaptive tool invocation proficiency.

\item A systematic analysis of critical open challenges and research directions, focusing on expanding retrieval scope beyond traditional methods, enabling asynchronous parallel execution, developing comprehensive multi-modal benchmarks, and optimising multi-agent architectures for enhanced robustness and efficiency.

\end{itemize}

\paragraph{Survey Organization.}

This survey methodically explores recent advancements in DR agents, organised as follows: Section 2 provides foundational concepts, examining recent progress in reasoning, retrieval-augmented generation, and agent communication protocols. Section 3 comprehensively analyses key DR agent components, including search engine integration (Section 3.1), tool invocation strategies (Section 3.2), architectural workflows (Section 3.3), and optimisation methodologies (Section 3.4). Section 4 reviews major industrial applications and practical implementations of DR agents by leading organisations. Section 5 surveys benchmarks used for evaluating DR systems, categorising them into question-answering and task execution scenarios. Section 6 highlights critical challenges and outlines promising directions for future research, focusing on enhancing information acquisition, asynchronous parallel execution, benchmark alignment, and optimising multi-agent architectures. Finally, Section 7 concludes with a summary and provides insights into the broader implications and opportunities within DR agent research.

\section{Background and Preliminaries}

\subsection{Advances in Reasoning and Tool Integration}

Recent advancements in large reasoning models (LRMs) have greatly enhanced the ability of language models to tackle complex and abstract tasks. These models have shown significant improvements in tasks such as arithmetic, common-sense reasoning, and symbolic problem-solving, largely due to innovations in model architectures and training techniques. One such advancement is Chain-of-Thought (CoT) prompting, introduced by Wei et al. \citep{wei2023chainofthoughtpromptingelicitsreasoning}, which explicitly guides models to articulate intermediate logical steps, decomposing complex problems into simpler, sequential stages. This has led to notable improvements in both the interpretability and accuracy of LLMs on various reasoning benchmarks. Building upon CoT, subsequent research has introduced methods to further enhance LLM reasoning, particularly in handling lengthy textual contexts. Approaches such as positional interpolation and sparse attention mechanisms \citep{bai2024longalign, wang2024beyond} have been proposed to extend the effective context window. Furthermore, specialised benchmarks like LongBench \citep{bai2024longbench} and LongFinanceQA \citep{lin2025facilitating} have been developed to rigorously evaluate and improve the performance of these models in extended-context reasoning.

To address reasoning tasks that require real-time or specialised external knowledge, frameworks like Toolformer \citep{schick2023toolformerlanguagemodelsteach} and MultiTool-CoT \citep{inaba2023multitool} have been proposed, enabling LLMs to autonomously incorporate external computational resources and APIs directly within reasoning workflows. These approaches effectively enhance performance in tasks dependent on precise numerical calculations and dynamic information retrieval. Maintaining reasoning coherence across multiple conversational turns also poses distinct challenges. Techniques such as Dialogue CoT \citep{chae2023dialogue} and Structured CoT (SCoT) \citep{sultan2024structured} explicitly integrate dialogue states and conversational contexts within reasoning chains, significantly improving coherence, context-awareness, and the ability to manage iterative interactions and clarify complex user queries. However, despite substantial improvements, existing reasoning frameworks still encounter critical issues, including hallucinations, static or outdated internal knowledge, and insufficient responsiveness to rapidly changing information needs. These limitations highlight the necessity of integrating external information sources, real-time retrieval mechanisms, and adaptive reasoning strategies, which are core motivations driving recent advances toward more comprehensive and robust reasoning frameworks suitable for DR Agent applications.

\subsection{Advances in Retrieval-Augmented Generation and Agentic Retrieval}
Retrieval-augmented Generation (RAG), leveraging external knowledge bases (e.g., webs, APIs), has emerged as an effective strategy to mitigate hallucination problems and enhance the accuracy of web information search \citep{fan2024survey,gao2023retrieval,singh2025agentic}. Early RAG architectures typically involved a static pipeline, where retrievers fetched relevant documents from external sources such as Wikipedia or search engines, and generators (e.g., LLMs) produced answers based solely on these retrieved passages. However, static approaches were limited in handling complex or multi-step queries, motivating recent advances toward iterative and interactive retrieval mechanisms to generate richer and more relevant responses, including FLARE \citep{zhang2024enhancing}, Self-RAG \citep{asai2023self}, IAG \citep{zhang2023iag}, and ToC \citep{kim2023tree}. In addition, studies \citep{izacard2023atlas, lin2023ra} expanded retrieval sources from structured databases (e.g., Wikipedia) to large-scale, diverse web corpora such as the Common Crawl dump preprocessed via the CCNet pipeline \citep{fu2022ccnet}. Further improvements of RAG include hybrid approaches that combine internal LLM knowledge and external retrievals for better accuracy and coherence \citep{aliannejadi2024trec}. Recently, Huang et al. \citep{huang2025rag} proposed RAG-RL, introducing reinforcement learning and curriculum learning techniques, enabling reasoning language models (RLMs) to more effectively identify and utilise relevant contexts.

Despite these advancements in retrieval methods and reasoning-enhanced models, RAG approaches still face limitations in effectively managing complex reasoning workflows and dynamically adapting to varied task requirements. To address these challenges, recent research extends RAG into an agentic paradigm, integrating additional reasoning and decision-making layers atop conventional RAG pipelines \citep{singh2025agentic}. Agentic RAG approaches leverage iterative retrieval, adaptive querying, and dynamic workflow adjustments, significantly enhancing multi-step reasoning capabilities. For example, RL-based query refinement techniques (e.g., Hsu et al. \citep{hsu2024grounding}) improve retrieval for complex queries, while graph-based retrieval (e.g., GeAR \citep{shen2024gear}) further enhances the processing of multi-hop queries. Despite these advancements, agentic RAG still faces critical challenges, including balancing computational overhead from dynamic reasoning processes \citep{singh2025agentic}, aligning agent behaviours with user intentions \citep{zerhoudi2024personarag}, and ensuring interpretability in adaptive workflows \citep{hsu2024grounding,singh2025agentic}. Moreover, even advanced agentic RAG approaches remain constrained by their reliance on pre-existing or periodically updated corpora, limiting their ability to handle real-time, rapidly changing, or long-tail information needs effectively. Addressing this challenge requires integrating external APIs and web browsing capabilities into RAG architectures, motivating recent DR methods aimed at further enhancing retrieval comprehensiveness and adaptability.

\subsection{Model Context Protocol and Agent-to-Agent Policy}
\vspace{-1mm}
Model Context Protocol (MCP) and Agent-to-Agent (A2A) have been proposed to address interoperability challenges in LLM-based agent systems, enabling efficient tool access and effective multi-agent collaboration. \textbf{MCP}: Traditional Tool Use (TU) agents face significant challenges, including inconsistent APIs, high maintenance costs, and redundant development efforts, severely limiting interoperability across systems \citep{schick2023toolformerlanguagemodelsteach}. To address these issues, Anthropic introduced the MCP, a unified communication layer allowing LLM-based agents to interact securely and consistently with external services and data sources via standardised interfaces. MCP mitigates data silo problems by providing dynamic service discovery and uniform access patterns. \textbf{A2A}: Google's A2A protocol facilitates decentralised multi-agent collaboration through structured, task-oriented dialogues. Agents from diverse vendors and model architectures can discover peers, delegate responsibilities, and collaboratively manage complex tasks as equal participants \citep{google2025a2a}. By abstracting agent discovery into Agent Cards and task coordination into Tasks and Artefacts, A2A supports flexible, incremental, multi-modal workflows, ideally suited to sophisticated collaborative scenarios.

MCP and A2A complement each other by clearly separating responsibilities: MCP serves as a standardised interface for accessing external tools, while A2A orchestrates collaborative agent interactions. Together, they establish a modular and scalable foundation for open, interoperable agent ecosystems, significantly enhancing the practical capabilities of AI systems in tackling complex real-world challenges.

\vspace{-2mm}
\section{Deep Research: Search Engine, Tool Use, Workflow, Tuning, Non-parametric Continual Learning}
\vspace{-2mm}
\textbf{Comparison with Conventional RAG-based Approaches.} DR agents expand the capabilities of traditional RAG methods by integrating dynamic retrieval, real-time TU, and adaptive reasoning into a unified system. RAG-based approaches typically rely on fixed pipelines, limiting their flexibility in handling complex, multi-step queries or rapidly changing contexts. In contrast, DR agents provide greater autonomy, context-awareness, and accuracy by dynamically engaging with external tools and managing multi-stage research tasks in real time.

In this section, we explore five core components essential for the development and optimization of DR agents: (3.1) \textbf{search engine integration}, which compares API-based interfaces with browser-based exploration to enhance dynamic knowledge acquisition; (3.2) \textbf{Tool Use capabilities}, which investigate the integration of code execution, mathematical computation, file manipulation, and multimodal processing modules within the agent’s inference pipeline; (3.3) \textbf{workflow architecture}, analysing foundational designs, the balance between multi-agent and single-agent paradigms, memory mechanisms, and auxiliary components that facilitate the orchestration of complex research workflows; (3.4) \textbf{tuning methodologies}, which examine prompt-driven structured generation, LLM-driven prompting, fine-tuning strategies, and reinforcement learning approaches aimed at optimizing agent performance, and (3.5) \textbf{Non-parametric continual learning}, which enables LLM agents to self-evolve by dynamically adapting external tools, memory, and workflows without updating internal model weights, offering scalable optimization for complex tasks.

\vspace{-1mm}
\begin{figure}[!htbp]
    \centering
    \includegraphics[width=\linewidth]{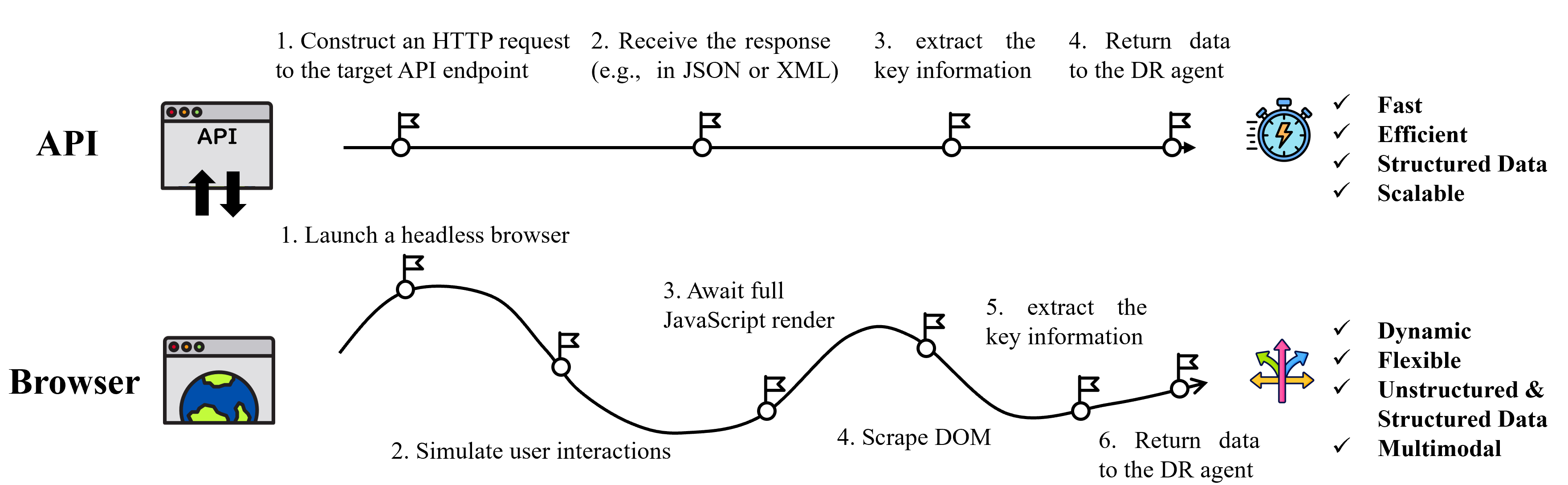}
    \caption{General Comparison of API-Based and Browser-Based Retrieval Workflow.}
    \label{fig31}
\end{figure}
\vspace{-2mm}
\subsection{Search Engine: API vs. Browser}
\vspace{-2mm}
To enhance reasoning depth and accuracy for handling evolving tasks, DR agents employ search engines (SE) to update their knowledge through interaction with the external environment. In Table \ref{tab:search_engine}, we present a comparative overview of SEs, base models, and evaluation benchmarks employed by existing DR agents. The SEs can be broadly categorised into two types:
\begin{enumerate}[label=\arabic*),itemsep=0pt,topsep=0pt,parsep=0pt,partopsep=0pt]
    \item \textbf{API-Based SEs}, which interact with structured data sources, such as search-engine APIs or scientific database APIs, enabling efficient retrieval of organised information.
    \item \textbf{Browser-Based SEs}, which simulate human-like interactions with web pages, facilitating real-time extraction of dynamic or unstructured content, improving the comprehensiveness of the external knowledge.
\end{enumerate}

\setlength{\tabcolsep}{8pt}

\begin{longtable}{%
  >{\arraybackslash}m{2.7cm}
  >{\centering\arraybackslash}m{0.3cm}
  >{\centering\arraybackslash}m{1.1cm}
  >{\centering\arraybackslash}m{0.4cm}
  >{\centering\arraybackslash}m{0.4cm}
  >{\centering\arraybackslash}m{3.3cm}
  >{\centering\arraybackslash}m{2.7cm}
  >{\centering\arraybackslash}m{1.3cm}
}
\caption{Comparison of DR Agents with Search Engine Details}
\label{tab:search_engine} \\

\multicolumn{8}{l}{
    \blackbox{} = \textbf{Primary focus},\ 
    \graybox{} = \textbf{Secondary/minor focus},\ 
    \whitebox{} = \textbf{Not present}
} \\
\toprule
\multirow{2}{*}{\textbf{DR Agent}} & \multicolumn{2}{c}{\textbf{Search Engine}}
& \multicolumn{3}{c}{\textbf{Benchmark}} & \multirow{2}{*}{\textbf{Base Model}} & \multirow{2}{*}{\textbf{Release}} \\
& \textbf{API} & \textbf{Browser} & \textbf{GAIA} & \textbf{HLE} & \textbf{Other QA} & & \\
\midrule
\endfirsthead

\multicolumn{8}{l}{\textit{Table \ref{tab:search_engine} continued from previous page}} \\
\toprule
\multirow{2}{*}{\textbf{DR Agent}} & \multicolumn{2}{c}{\textbf{Search Engine}}
& \multicolumn{3}{c}{\textbf{Benchmark}} & \multirow{2}{*}{\textbf{Base Model}} & \multirow{2}{*}{\textbf{Release}} \\
& \textbf{API} & \textbf{Browser} & \textbf{GAIA} & \textbf{HLE} & \textbf{Other QA} & & \\
\midrule
\endhead

\midrule
\multicolumn{8}{r}{\textit{Continued on next page}} \\
\endfoot

\bottomrule
\endlastfoot

Avatar \cite{wu2024avataroptimizingllmagents} & \graybox & \whitebox & \whitebox & \whitebox & Stark & Claude-3-Opus, GPT-4 & \href{https://arxiv.org/abs/2406.11200}{Feb-2024} \\
\rowcolor{gray!10}
CoSearch-Agent \cite{gong2024cosearchagent} & \blackbox & \whitebox & \whitebox & \whitebox & \whitebox & GPT-3.5-turbo & \href{https://arxiv.org/abs/2402.06360}{Feb-2024} \\
MMAC-Copilot \cite{song2024mmac} & \blackbox & \whitebox & \blackbox & \whitebox & \whitebox & GPT-3.5, GPT-4 & \href{https://arxiv.org/abs/2404.18074}{Mar-2024} \\
\rowcolor{gray!10}
Storm \cite{shao2024assistingwritingwikipedialikearticles} & \graybox & \whitebox & \whitebox & \whitebox & FreshWiki & GPT-3.5-turbo & \href{https://arxiv.org/abs/2402.14207}{Jul-2024} \\
OpenResearcher \cite{zheng2024openresearcher} & \blackbox & \whitebox & \whitebox & \whitebox & Privately Collected QA Data & DeepSeek-V2-Chat & \href{https://arxiv.org/abs/2408.06941}{Aug-2024} \\
\rowcolor{gray!10}
The AI Scientist \cite{lu2024aiscientistfullyautomated} & \blackbox & \whitebox & \whitebox & \whitebox & MLE-Bench & GPT-4o, o1-mini, o1-preview & \href{https://arxiv.org/abs/2408.06292}{Aug-2024} \\
Gemini DR \cite{geminideepresearch} & \blackbox & \blackbox & \whitebox & \blackbox & GPQA & Gemini-2.0-Flash & \href{https://gemini.google/overview/deep-research/}{Dec-2024} \\
\rowcolor{gray!10}
Agent Laboratory \cite{schmidgall2025agent} & \blackbox & \whitebox & \whitebox & \whitebox & MLE-Bench & GPT-4o, o1-preview & \href{https://arxiv.org/abs/2501.04227}{Jan-2025} \\
Search-o1 \cite{li2025search} & \blackbox & \whitebox & \whitebox & \whitebox & GPQA, NQ, TriviaQA & QwQ-32B-preview & \href{https://arxiv.org/abs/2501.05366}{Jan-2025} \\
\rowcolor{gray!10}
WebWalker \cite{wu2025webwalker} & \whitebox & \whitebox & \whitebox & \whitebox & WebWalkerQA & GPT-4o, Qwen-2.5 & \href{https://arxiv.org/abs/2501.07572}{Jan-2025} \\
Agentic Reasoning \cite{wu2025agentic} & \blackbox & \whitebox & \whitebox & \whitebox & GPQA & DeepSeek-R1, Qwen2.5 & \href{https://arxiv.org/abs/2502.04644}{Feb-2025} \\
\rowcolor{gray!10}
AutoAgent \cite{tang2025autoagentfullyautomatedzerocodeframework} & \whitebox & \blackbox & \blackbox & \whitebox & \whitebox & Claude-Sonnet-3.5 & \href{https://arxiv.org/abs/2502.05957}{Feb-2025} \\
Grok DeepSearch \cite{grokdeepresearch} & \blackbox & \blackbox & \whitebox & \whitebox & GPQA & Grok3 & \href{https://x.ai/news/grok-3}{Feb-2025} \\
\rowcolor{gray!10}
OpenAI DR \cite{openai2025deepresearch} & \whitebox & \blackbox & \blackbox & \blackbox & \blackbox & GPT-o3 & \href{https://openai.com/index/introducing-deep-research/}{Feb-2025} \\
Perplexity DR \cite{perplexitydeepresearch} & \blackbox & \graybox & \whitebox & \blackbox & SimoleQA & Flexible & \href{https://www.perplexity.ai/hub/blog/introducing-perplexity-deep-research}{Feb-2025} \\
\rowcolor{gray!10}
Towards an AI co-scientist \cite{gottweis2025towards} & \blackbox & \whitebox & \whitebox & \whitebox & GPQA & Gemini 2.0 & \href{https://arxiv.org/abs/2502.18864}{Feb-2025} \\
Nouswise \cite{nouswise2025} & \whitebox & \whitebox & \whitebox & \whitebox & \whitebox & — & \href{https://nouswise.com/homepage}{Mar-2025} \\
\rowcolor{gray!10}
AgentRxiv \cite{schmidgall2025agentrxiv} & \blackbox & \whitebox & \whitebox & \whitebox & GPQA, MedQA & GPT-4o-mini & \href{https://arxiv.org/abs/2503.18102}{Mar-2025} \\
Agent-R1 \cite{Agent-R1} & \blackbox & \whitebox & \whitebox & \whitebox & HotpotQA & Qwen2.5-1.5B-Inst & \href{https://github.com/0russwest0/Agent-R1}{Mar-2025} \\
\rowcolor{gray!10}
AutoGLM Rumination\cite{zhipu2025autoglm} & \whitebox & \blackbox & \whitebox & \whitebox & GPQA & GLM-Z1-Air & \href{https://autoglm-research.zhipuai.cn/}{Mar-2025} \\
Copilot Researcher \cite{microsoft_copilot_researcher} & \whitebox & \blackbox & \whitebox & \whitebox & \whitebox & o3-mini & \href{https://www.microsoft.com/en-us/microsoft-365/blog/2025/03/25/introducing-researcher-and-analyst-in-microsoft-365-copilot/}{Mar-2025} \\
\rowcolor{gray!10}
H2O.ai DR \cite{h2oai} & \blackbox & \blackbox & \blackbox & \whitebox & \whitebox & h2ogpt-oasst1-512-12b & \href{https://h2o.ai/}{Mar-2025} \\
Manus \cite{manus2025} & \blackbox & \blackbox & \whitebox & \whitebox & \whitebox & Claude3.5, GPT-4o & \href{https://manus.im/}{Mar-2025} \\
\rowcolor{gray!10}
Openmanus \cite{openmanus2025} & \blackbox & \blackbox & \whitebox & \whitebox & \whitebox & Claude3.5, GPT-4o & \href{https://github.com/mannaandpoem/OpenManus}{Mar-2025} \\
OWL \cite{owl2025} & \blackbox & \blackbox & \blackbox & \whitebox & \whitebox & Deepeek-R1, Gemini2.5-Pro, GPT-4o & \href{https://github.com/camel-ai/owl}{Mar-2025} \\
\rowcolor{gray!10}
R1-Searcher \cite{song2025r1} & \graybox & \whitebox & \whitebox & \whitebox & 2WikiMultiHopQA, HotpotQA & Llama3.1-8B-Inst, Qwen2.5-7B & \href{https://arxiv.org/abs/2503.05592}{Mar-2025} \\
ReSearch \cite{chen2025learning} & \graybox & \whitebox & \whitebox & \whitebox & 2WikiMultiHopQA, HotpotQA & Qwen2.5-7B, Qwen2.5-7B-Inst & \href{https://arxiv.org/abs/2503.19470}{Mar-2025} \\
\rowcolor{gray!10}
Search-R1 \cite{jin2025search} & \blackbox & \blackbox & \whitebox & \whitebox & 2WikiMultiHopQA, HotpotQA, NQ, TriviaQA & Llama3.2-3B, Qwen2.5-3B/7B & \href{https://arxiv.org/abs/2503.09516}{Mar-2025} \\
DeepResarcher \cite{zheng2025deepresearcherscalingdeepresearch} & \whitebox & \blackbox & \blackbox & \whitebox & HotpotQA, NQ, TriviaQA & Qwen2.5-7B-Inst & \href{https://arxiv.org/abs/2504.03160}{Apr-2025} \\
\rowcolor{gray!10}
Genspark Super Agent \cite{genspark} & \blackbox & \blackbox & \blackbox & \whitebox & \whitebox & Mixture of Agents\tablefootnote{Mixture of Agents refers to an ensemble of nine base models comprising GPT-4.1, GPT-o3, GPT-o4-mini-high, Claude-Sonnet-3.7-Thinking, Claude-Sonnet-3.7, Gemini-2.0-Flash, Gemini-2.5-Pro, DeepSeek-V3, DeepSeek-R1} & \href{https://www.genspark.ai/}{Apr-2025} \\
WebThinker \cite{Li2025webthinker} & \blackbox & \blackbox & \blackbox & \blackbox & GPQA, WebWalkerQA & QwQ-32B & \href{https://github.com/RUC-NLPIR/WebThinker}{Apr-2025} \\
\rowcolor{gray!10}
SWIRL \cite{goldie2025synthetic} & \blackbox & \whitebox & \whitebox & \whitebox & HotQA, BeerQA & Gemma 2-27b & \href{https://arxiv.org/abs/2504.04736}{Apr-2025} \\
SimpleDeepSearcher \cite{SimpleDeepSearcher} & \whitebox & \blackbox & \blackbox & \whitebox & 2WikiMultiHopQA & Qwen-2.5-7B-In, Qwen-2.5-32B-In, DeepSeek-Distilled-Qwen-2.5-32B, QwQ-32B & \href{https://github.com/RUCAIBox/SimpleDeepSearcher}{Apr-2025} \\
\rowcolor{gray!10}
Suna AI \cite{sunaai} & \blackbox & \blackbox & \whitebox & \whitebox & \whitebox & GPT-4o, Claude & \href{https://github.com/kortix-ai/suna}{Apr-2025} \\
Tool-Star \cite{dong2025tool}& \blackbox & \blackbox & \blackbox & \blackbox & WebWalker, HotpotQA, 2WikiMultiHopQA & Qwen-2.5 & \href{https://arxiv.org/abs/2505.16410}{May-2025} \\
\rowcolor{gray!10}
WebDancer \cite{wu2025webdancer} & \blackbox & \blackbox & \blackbox & \whitebox & WebWalkerQA & Qwen-2.5, QwQ-32B, DeepSeek-R1, GPT-4o & \href{https://arxiv.org/abs/2505.22648}{May-2025} \\
AgenticSeek \cite{agenticseek} & \whitebox & \blackbox & \whitebox & \whitebox & \whitebox & GPT-4o, DeepSeek-R1, Claude & \href{https://github.com/Fosowl/agenticSeek}{May-2025} \\
\rowcolor{gray!10}
Alita \cite{qiu2025alita} & \blackbox & \blackbox & \blackbox & \whitebox & PathVQA & GPT-4o, Claude-Sonnet-4 & \href{https://arxiv.org/pdf/2505.20286}{May-2025} \\
DeerFlow \cite{deerflow} & \blackbox & \whitebox & \whitebox & \whitebox & \whitebox & Doubao-1.5-Pro-32k, DeepSeek-R1, GPT-4o, Qwen & \href{https://github.com/bytedance/deer-flow}{May-2025} \\
\rowcolor{gray!10}
PANGU DEEPDIVER \cite{shi2025pangu} & \blackbox & \whitebox & \whitebox & \whitebox & C-SimpleQA, HotpotQA, ProxyQA & Pangu-7B-Reasoner & \href{https://arxiv.org/pdf/2505.24332}{May-2025} \\
O-agents \cite{zhu2025oagents} & \blackbox & \whitebox & \blackbox & \whitebox & \whitebox & GPT-4o, GPT-4.1, Claude-3.7-Sonnet, DeepSeek-R1, Gemini-2.5 & \href{https://arxiv.org/abs/2506.15741}{Jun-2025} \\
\rowcolor{gray!10}
Kimi-Researcher \cite{kimi2025researcher} & \blackbox & \blackbox & \whitebox & \blackbox & SimpleQA & Kimi k1.5/k2 & \href{https://moonshotai.github.io/Kimi-Researcher/}{Jun-2025} \\
WebSailor \cite{li2025websailor} & \blackbox & \blackbox & \blackbox & \whitebox & SimpleQA & Qwen-2.5 & \href{https://arxiv.org/abs/2507.02592}{Jul-2025} \\
\rowcolor{gray!10}
Agent-KB \cite{tang2025agent} & \blackbox & \whitebox & \blackbox & \whitebox & SWE-bench & GPT-4o, GPT-4.1, Claude-3.7-Sonnet, o3-mini, Qwen-3, DeepSeek-R1 & \href{https://arxiv.org/abs/2507.06229}{Jul-2025} \\
WebShaper \cite{tao2025webshaper} & \blackbox & \blackbox & \blackbox & \whitebox & WebWalkerQA & Qwen-2.5, QwQ-32B & \href{https://arxiv.org/abs/2507.15061}{Jul-2025} \\
\rowcolor{gray!10}
Deep Researcher with Test-Time Diffusion \cite{han2025deep} & \blackbox & \whitebox & \blackbox & \blackbox & \whitebox & Gemini-2.5-Pro & \href{https://arxiv.org/abs/2507.16075}{Jul-2025} \\
ChatGPT-Agent \cite{openai2025chatgptagent} & \whitebox & \whitebox & \whitebox & \whitebox & \whitebox & — & \href{https://openai.com/zh-Hans-CN/index/introducing-chatgpt-agent/}{Jul-2025} \\
\rowcolor{gray!10}
AWorld \cite{aworld2025} & \blackbox & \blackbox & \blackbox & \whitebox & HotpotQA & Gemini-2.5-Pro, GPT-4o & \href{https://github.com/inclusionAI/AWorld}{Jul-2025} \\
Cognitive Kernel-Pro \cite{wan2025cognitive} & \blackbox & \whitebox & \blackbox & \whitebox & AgentWebQA, WebWalkerQA, Multi-hop URLQA, DocBench, TableBench & Claude-3.7-Sonnet, CK-Pro-8B & \href{https://arxiv.org/abs/2508.00414}{Aug-2025} \\
\rowcolor{gray!10}
WebWatcher \cite{geng2025webwatcher} & \blackbox & \blackbox & \whitebox & \blackbox & Browsercom-VL, LiveVQA, MMSearch & Qwen-2.5-VL-32B & \href{https://arxiv.org/abs/2508.05748}{Aug-2025} \\
WideSearch \cite{wong2025widesearch} & \blackbox & \whitebox & \whitebox & \whitebox & WideSearch & DeepSeek-R1, Doubao-Seed-1.6, Claude Sonnet 4, Gemini-2.5-Pro & \href{https://arxiv.org/abs/2508.07999}{Aug-2025} \\
\rowcolor{gray!10}
MiroRL \cite{2025mirorl} & \blackbox & \blackbox & \blackbox & \whitebox & \whitebox & Qwen3-14B & \href{https://github.com/MiroMindAI/MiroRL}{Aug-2025} \\

\end{longtable}

API-based retrieval is a fast, efficient, and scalable way for deep-research (DR) agents to access external knowledge with relatively low latency and computational overhead. For example, Gemini DR~\citep{geminideepresearch} coordinates multiple interfaces, most notably the Google Search and arXiv APIs, to conduct large-scale retrieval over hundreds to thousands of web pages, substantially broadening coverage. Grok DeepSearch~\citep{grokdeepresearch} maintains a continuously updated index via news-outlet feeds, the Wikipedia API, and X’s native interface, and, on demand, dispatches a query-driven agent to decompose questions into targeted subqueries and fetch relevant pages in real time. \citet{perplexitydeepresearch} first crawls hundreds of sources and then aggregates them to produce a final report. Cognitive Kernel-Pro~\citep{wan2025cognitive} leverages the free DuckDuckGo search interface to enable a fully open-source, low-cost DR pipeline. Agentic Reasoning, ReSearch, R1-Search and SWIRL~\citep{wu2025agentic,chen2025learning,song2025r1,goldie2025synthetic} explicitly teach models when to search, what to search for, and how to incorporate retrieved evidence into the reasoning process. PANGU DeepDiver~\citep{shi2025pangu} uses reinforcement learning to adapt search intensity to task difficulty. Agent Laboratory~\citep{schmidgall2025agent} calls the arXiv API to extract paper metadata and abstracts for automated literature reviews, while AI Scientist~\citep{lu2024aiscientistfullyautomated} queries the Semantic Scholar API to validate novelty and citation relations among model-generated ideas. CoSearch-Agent~\citep{gong2024cosearchagent} integrates SerpApi to deliver Slack-based, real-time search. DeepRetrieval~\citep{jiang2025deepretrieval} operates within a reinforcement-learning framework to optimise queries against the PubMed and ClinicalTrials.gov APIs for high-recall biomedical retrieval, and Search-o1~\citep{li2025search} combines the Bing Search API with the Jina Reader API to dynamically extract and refine passages for downstream reasoning. While these API-driven approaches excel at structured, high-throughput acquisition, they can struggle with deeply nested, client-side JavaScript–rendered content, interactive components, or authentication barriers, motivating complementary browser-based mechanisms capable of comprehensively extracting and analysing dynamic or unstructured information.

Browser-based retrieval provides DR agents with dynamic, flexible, and interactive access to multimodal and unstructured web content through simulated human-like browser interactions. For example, Manus AI’s browsing agent operates a sandboxed Chromium instance for each research session, programmatically opening new tabs, issuing search queries, clicking through result links, scrolling pages until content thresholds are met, filling out form elements when necessary, executing in-page JavaScript to reveal lazily loaded sections, and downloading files or PDFs for local analysis \citep{manus2025}. Although OpenAI DR, Grok DeepSearch, and Gemini 2.5 DR do not publicly disclose the implementation details of their browsing capabilities, their ability to handle interactive widgets, dynamically rendered content, and multi-step navigation strongly suggests that they too employ comparable headless-browser frameworks behind the scenes. Among open-source studies, AutoAgent \citep{yu2024auto} operates within a BrowserGym environment to scroll, interact with page components, and download files when APIs are unavailable; DeepResearcher \citep{zheng2025deepresearcherscalingdeepresearch} employs a dedicated Web Browsing Agent that, upon receiving a browse request, processes each segment of a webpage in turn, decides whether to continue to subsequent segments based on relevance, and incrementally aggregates pertinent information into a short-term memory buffer before returning it for reasoning. Kimi-Researcher~\citep{kimi2025researcher} uses an internal search engine with a text-based browser to retrieve information. Search-R1 and MiroRL~\citep{jin2025search,2025mirorl} employ both search and browser tools during training. AutoGLM~\citep{zhipu2025autoglm} operationalises browsing through a plan–execute loop that opens and reads web pages, layering ``rumination'' cycles over browser actions to refine evidence and produce long-form reports. Genspark Super Agent~\citep{genspark} orchestrates a mixture-of-agents in which a research sub-agent performs web search and page reading, handing structured notes to downstream writing/analysis agents rather than relying on a single monolithic browser. SimpleDeepSearcher~\citep{SimpleDeepSearcher} follows a lightweight search-fetch-summarise loop using web search APIs plus HTTP fetching in lieu of full browser automation, caching pages and compressing them before further reasoning. Tool-Star~\citep{dong2025tool} explicitly separates a Search Engine tool from a Web Browser Agent: after link retrieval, the browser agent opens pages, extracts salient snippets, and returns compressed evidence to the planner. AgenticSeek~\citep{agenticseek} couples a local metasearch front end with a headless, stealth browser so the agent can click, scroll, and submit forms on live sites, exposing knobs for budget and anti-bot robustness. AWorld~\citep{aworld2025} provides a multi-agent runtime with built-in browser automation and tracing, enabling teams of agents to divide research, browsing, and synthesis for deep-research workflows over dynamic sites. WebThinker~\citep{Li2025webthinker} performs information seeking by issuing searches and following links on returned result pages. WebDancer, WebSailor, and WebShaper~\citep{wu2025webdancer,li2025websailor,tao2025webshaper} combine web search with on-page navigation as a minimal sufficient toolset, linking structured tool invocations with observations in a ReAct-style closed loop: they first locate candidate sources horizontally, then drill down vertically, trading a compact action space for training stability and stronger generalisation. WebWatcher~\citep{geng2025webwatcher} uses the Google SerpApi for multimodal search and applies OCR-based image processing to emulate browser interactions. While browser-based retrieval excels at capturing real-time and deeply nested content that API calls cannot reach, it also incurs greater latency, resource consumption, and complexity in handling page variability and errors, suggesting that DR agents may benefit from hybrid architectures that combine the efficiency of API-based methods with the comprehensiveness of browser-driven exploration.

\subsection{Tool Use: Empowering Agents with Extended Functionalities}

\begin{table}[!htbp]
  {\scriptsize
  \setlength{\tabcolsep}{4.5pt}
  \rowcolors{2}{gray!10}{white}
  \centering
  \caption{Comparison of DR Agents with Tool Use Capabilities}
  \label{tab:tool_use}
  \begin{tabular}{%
    >{\centering\arraybackslash}m{3.6cm}
    >{\centering\arraybackslash}m{3.1cm}
    >{\centering\arraybackslash}m{3.1cm}
    >{\centering\arraybackslash}m{3.1cm}
    >{\centering\arraybackslash}m{2.0cm}
  }
    \multicolumn{5}{l}{\small
      \blackbox{} = \textbf{Involved},\ 
      \graybox{} = \textbf{Non Disclosure},\
      \whitebox{} = \textbf{Not present}
    }\\
    \toprule
    {\small \textbf{DR Agent}} & {\small \textbf{Code Interpreter}}
    & {\small \textbf{Data Analytics}} & {\small \textbf{Multimodal}} & {\small \textbf{Release}} \\
    \midrule
    CoSearchAgent \cite{gong2024cosearchagent} & \whitebox & \blackbox & \whitebox & \href{https://arxiv.org/abs/2402.06360}{Feb-2024} \\
    Storm \cite{shao2024assistingwritingwikipedialikearticles} & \blackbox & \whitebox & \whitebox & \href{https://arxiv.org/abs/2402.14207}{Jul-2024} \\
    The AI Scientist \cite{lu2024aiscientistfullyautomated} & \blackbox & \whitebox & \whitebox & \href{https://arxiv.org/abs/2408.06292}{Aug-2024} \\
    Agent Laboratory \cite{schmidgall2025agent} & \blackbox & \whitebox & \whitebox & \href{https://arxiv.org/abs/2501.04227}{Jan-2025} \\
    Agentic Reasoning \cite{wu2025agentic} & \blackbox & \whitebox & \whitebox & \href{https://arxiv.org/abs/2502.04644}{Feb-2025} \\
    AutoAgent \cite{tang2025autoagentfullyautomatedzerocodeframework} & \blackbox & \whitebox & \blackbox & \href{https://arxiv.org/abs/2502.05957}{Feb-2025} \\
    Genspark DR \cite{genspark} & \blackbox & \blackbox & \blackbox & \href{https://www.genspark.ai/}{Feb-2025} \\
    Grok DeepSearch \cite{grokdeepresearch} & \blackbox & \blackbox & \blackbox & \href{https://x.ai/news/grok-3}{Feb-2025} \\
    OpenAI DR \cite{openai2025deepresearch} & \blackbox & \blackbox & \blackbox & \href{https://openai.com/index/introducing-deep-research/}{Feb-2025} \\
    Perplexity DR \cite{perplexitydeepresearch} & \blackbox & \blackbox & \blackbox & \href{https://www.perplexity.ai/hub/blog/introducing-perplexity-deep-research}{Feb-2025} \\
    Towards an AI co-scientist \citep{gottweis2025towards} & \whitebox & \blackbox & \blackbox & \href{https://arxiv.org/abs/2502.18864}{Feb-2025} \\
    Agent-R1 \cite{Agent-R1} & \blackbox & \whitebox & \whitebox & \href{https://github.com/0russwest0/Agent-R1}{Mar-2025} \\
    AutoGLM Romination \cite{zhipu2025autoglm} & \blackbox & \whitebox & \blackbox & \href{https://autoglm-research.zhipuai.cn/}{Mar-2025} \\
    Copilot Researcher \cite{microsoft_copilot_researcher} & \blackbox & \blackbox & \graybox & \href{https://www.microsoft.com/en-us/microsoft-365/blog/2025/03/25/introducing-researcher-and-analyst-in-microsoft-365-copilot/}{Mar-2025} \\
    Manus \cite{manus2025} & \blackbox & \blackbox & \blackbox & \href{https://manus.im/}{Mar-2025} \\
    OpenManus \cite{openmanus2025} & \blackbox & \blackbox & \whitebox & \href{https://github.com/mannaandpoem/OpenManus}{Mar-2025} \\
    OWL \cite{owl2025} & \blackbox & \blackbox & \blackbox & \href{https://github.com/camel-ai/owl}{Mar-2025} \\
    H2O.ai DR \cite{h2oai} & \blackbox & \blackbox & \blackbox & \href{https://h2o.ai/}{Mar-2025} \\
    Genspark Super Agent \cite{genspark}  & \blackbox & \blackbox & \blackbox & \href{https://www.genspark.ai/}{Apr-2025} \\
    WebThinker \cite{Li2025webthinker} & \blackbox & \blackbox & \whitebox & \href{https://github.com/RUC-NLPIR/WebThinker}{Apr-2025} \\
    Suna Ai \cite{sunaai} & \blackbox & \blackbox & \whitebox & \href{https://github.com/kortix-ai/suna}{Apr-2025} \\
    Tool-Star \citep{dong2025tool} & \blackbox & \blackbox & \whitebox & \href{https://arxiv.org/abs/2505.16410}{May-2025} \\
    AgenticSeek \cite{agenticseek} & \blackbox & \blackbox & \whitebox & \href{https://github.com/Fosowl/agenticSeek}{May-2025} \\
    Alita \cite{qiu2025alita} & \blackbox & \graybox & \graybox & \href{https://arxiv.org/pdf/2505.20286}{May-2025} \\
    DeerFlow \cite{deerflow} & \blackbox & \blackbox & \whitebox & \href{https://github.com/bytedance/deer-flow}{May-2025} \\
    O-agents \citep{zhu2025oagents} & \blackbox & \blackbox & \blackbox & \href{https://arxiv.org/abs/2506.15741}{Jun-2025} \\
    Kimi-Researcher \citep{kimi2025researcher} & \blackbox & \blackbox & \whitebox & \href{https://moonshotai.github.io/Kimi-Researcher/}{Jun-2025} \\
    Agent-KB \citep{tang2025agent} & \blackbox & \blackbox & \blackbox & \href{https://arxiv.org/abs/2507.06229}{Jul-2025} \\
    AWorld \citep{aworld2025} & \blackbox & \blackbox & \blackbox & \href{https://github.com/inclusionAI/AWorld}{Jul-2025} \\
    Cognitive Kernel-Pro \citep{wan2025cognitive} & \blackbox & \blackbox & \blackbox & \href{https://arxiv.org/abs/2508.00414}{Aug-2025} \\
    WebWatcher \citep{geng2025webwatcher} & \blackbox & \blackbox & \blackbox & \href{https://arxiv.org/abs/2508.05748}{Aug-2025} \\
    MiroRL \citep{2025mirorl} & \blackbox & \blackbox & \whitebox & \href{https://github.com/MiroMindAI/MiroRL}{Aug-2025} \\
    \bottomrule
  \end{tabular}}
\end{table}

To expand DR agents’ capacity to interact with external environments in complex research tasks, specifically by actively invoking and handling diverse tools and data sources, various DR agents have introduced three core tool modules: code interpreters, data analytics, multimodal processing, along with the Model Context Protocol.

\paragraph{Code Interpreter.} 
The code interpreter capability enables DR agents to execute scripts during inference, allowing them to perform data processing, algorithm verification and model simulation. Most DR agents, except CoSearchAgent, embed a script execution environment. They typically rely on Python utilities such as Aider and Java utilities to orchestrate dynamic scripting, conduct literature-driven analysis and carry out real-time computational reasoning.

\paragraph{Data Analytics.}
By integrating data analytics modules, DR agents transform raw retrievals into structured insights by computing summary statistics, generating interactive visualisations and conducting quantitative model evaluations, thereby accelerating hypothesis testing and decision-making. Many commercial DR agents have implemented analytics features such as charting, table generation and statistical analysis, either locally or via remote services. However, most of these systems have not publicly disclosed technical details of their implementations. In contrast, academic studies often provide concrete examples: CoSearchAgent~\citep{gong2024cosearchagent} integrates SQL-based queries within team communication platforms to run aggregate analyses and produce reports; AutoGLM~\citep{zhipu2025autoglm} extracts and analyses structured datasets directly from table-based web interfaces.

\paragraph{Multimodal Processing and Generation.}
Multimodal processing and generation tools enable DR agents to integrate, analyse and generate heterogeneous data such as text, images, audio and video within a unified reasoning pipeline, thereby enriching their contextual understanding and broadening the range of their outputs. Only a subset of mature commercial and open-source projects, for example Manus~\citep{manus2025}, OWL~\citep{owl2025}, AutoAgent~\citep{tang2025autoagentfullyautomatedzerocodeframework}, AutoGLM~\citep{zhipu2025autoglm}, OpenAI~\citep{openai2025deepresearch}, Gemini~\citep{geminideepresearch}, Perplexity~\citep{perplexitydeepresearch} and Grok DeepSearch~\citep{grokdeepresearch}, support this capability, whereas most academic prototypes have not implemented it, often due to the high computational cost. As the typical open source studies, OWL and Openmanus extend their pipelines to include interactions with platforms such as GitHub, Notion and Google Maps and to leverage numerical libraries such as Sympy and Excel for combined data analysis and multimodal media processing \citep{owl2025,openmanus2025}.

\paragraph{Deep Research Agent with Computer Use.}
Most recently, the boundaries of DR agents have been progressively expanded through integrating computer-assisted task execution capabilities (i.e., computer use). For example, Zhipu AI introduced AutoGLM Rumination \citep{zhipu2025autoglm}, an RL-based system incorporating self-reflection and iterative refinement mechanisms, which significantly enhances multi-step reasoning and advanced function-calling abilities. Specifically, AutoGLM Rumination~\citep{zhipu2025autoglm} autonomously interacts with web environments, executes code, invokes external APIs, and effectively accomplishes sophisticated tasks, including data retrieval, analysis, and structured generation of comprehensive reports. Comparison with OpenAI's DR: While OpenAI DR primarily focus on intricate reasoning and information retrieval, AutoGLM Rumination exhibits superior autonomy in practical execution. This enhanced autonomy allows it to transform abstract analytical insights into concrete operational tasks, such as automated interactions with web interfaces and real-time data processing. Moreover, AutoGLM Rumination addresses and resolves limitations inherent in simulated browsing environments by seamlessly integrating advanced reasoning capabilities with authentic browser-based interactions. Therefore, the agent gains reliable access to user-authenticated resources, including platforms such as CNKI, Xiaohongshu, and WeChat official accounts. Such integration significantly elevates the agent’s autonomy and adaptability in both information acquisition and execution of real-world tasks.

OpenAI DR, Perplexity DR, Grok DR, H2O, Manus, Genspark Super Agent~\citet{openai2025deepresearch,perplexitydeepresearch,grokdeepresearch,h2oai,manus2025,genspark} orchestrate multi-step web research with browser or tool use, code execution, and analytics to produce cited, structured reports at industrial scale. OWL, OpenManus, Suna, DeerFlow, WebThinker, AgenticSeek, and AWorld~\citep{owl2025,openmanus2025,sunaai,deerflow,Li2025webthinker,agenticseek,aworld2025} provide open-source stacks that integrate browser automation, code interpreters, and (often) MCP-style tooling for end-to-end deep research workflows. AutoGLM Rumination, Tool-Star, Kimi-Researcher, and MiroRL~\citep{zhipu2025autoglm,dong2025tool,kimi2025researcher,2025mirorl} use reinforcement learning or self-reflection to plan searches, invoke tools and code, and improve multi-step reasoning autonomy. The AI Scientist, Storm, Agent Laboratory, Agent-R1, AutoAgent, and CoSearchAgent~\citep{lu2024aiscientistfullyautomated,shao2024assistingwritingwikipedialikearticles,schmidgall2025agent,Agent-R1,tang2025autoagentfullyautomatedzerocodeframework,gong2024cosearchagent} automate literature review, code/experiment execution, and structured drafting through tool-augmented research pipelines. Towards an AI co-scientist and O-agents~\citep{gottweis2025towards,zhu2025oagents} outline blueprints and empirical recipes for building, training, and evaluating effective tool-using research agents. Agent-KB~\citep{tang2025agent} introduces a KB-driven framework for cross-domain experience transfer to boost generalisation on complex tasks. Microsoft Copilot Researcher~\citep{microsoft_copilot_researcher} embeds multi-step research and analytics in the Microsoft 365 ecosystem, producing reports and charts within enterprise workflows. Alita~\citep{qiu2025alita} explores self-evolving agents that generate and wrap MCP tools, alongside code execution, to extend capabilities with minimal predefined schemas.

\subsection{Architecture and Workflow}

As shown in Figure \ref{fig:workflow}, this section systematically analyses the construction of DR systems, focusing on workflows categorised into \textbf{static} and \textbf{dynamic} types. We first introduce the static workflows and then discuss planning strategies, which enhance task allocation and execution through \textbf{three distinctive user interaction types} to clarify intent: planning-only (direct planning without clarifying user intent), intent-to-planning (clarifying intent before planning to align the task with user goals), and unified intent-planning (generating a plan and requesting user confirmation). The distinction between \textbf{single-agent} and \textbf{multi-agent} systems is examined in the context of dynamic workflows, emphasising specialisation in task management. Additionally, we examine memory mechanisms for managing and integrating retrieved information, which enhance the performance and adaptability of DR systems.

\begin{figure}[htbp]
    \centering
    \includegraphics[width=\linewidth]{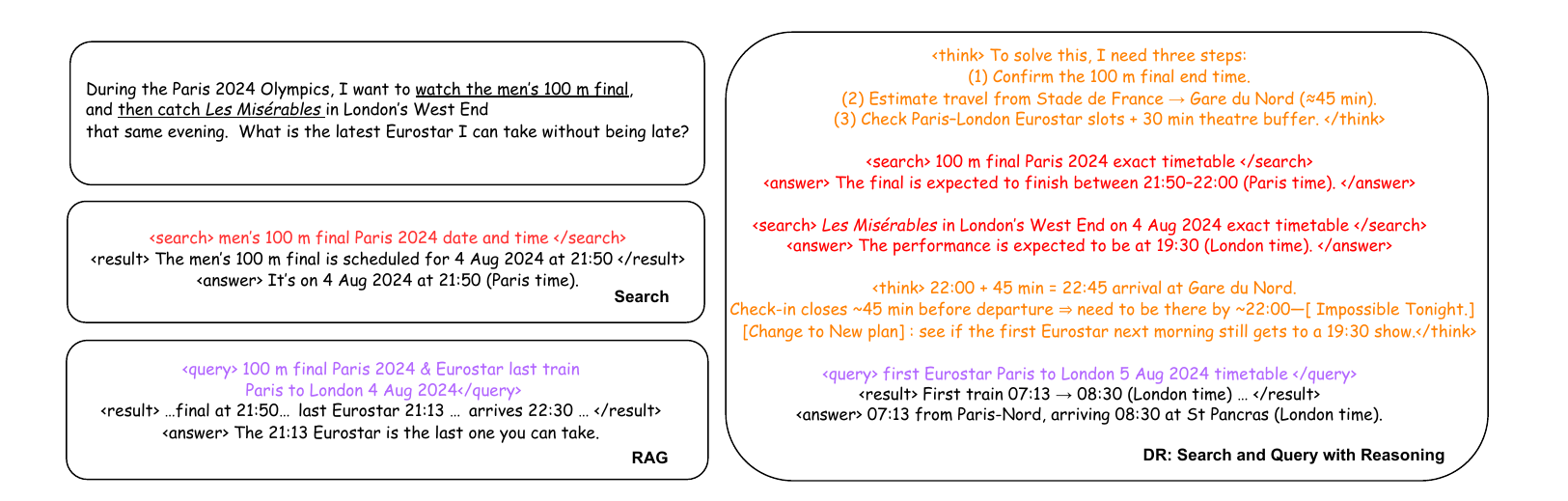}
    \caption{\textbf{Comparison of Information Retrieval Methods.} The upper left corner (Search) represents the searching methods, which can use the browser or API; the lower left corner (RAG, Query) represents Retrieval-Augmented Generation, combining retrieval and generative models to output natural language answers; the right side (Deep Research) represents the deep research process, generating complex decisions or analyses through retrieval and explicit reasoning.}
    \label{fig:information}
\end{figure}

\begin{figure}[!htbp]
    \centering
    \includegraphics[width=\linewidth]{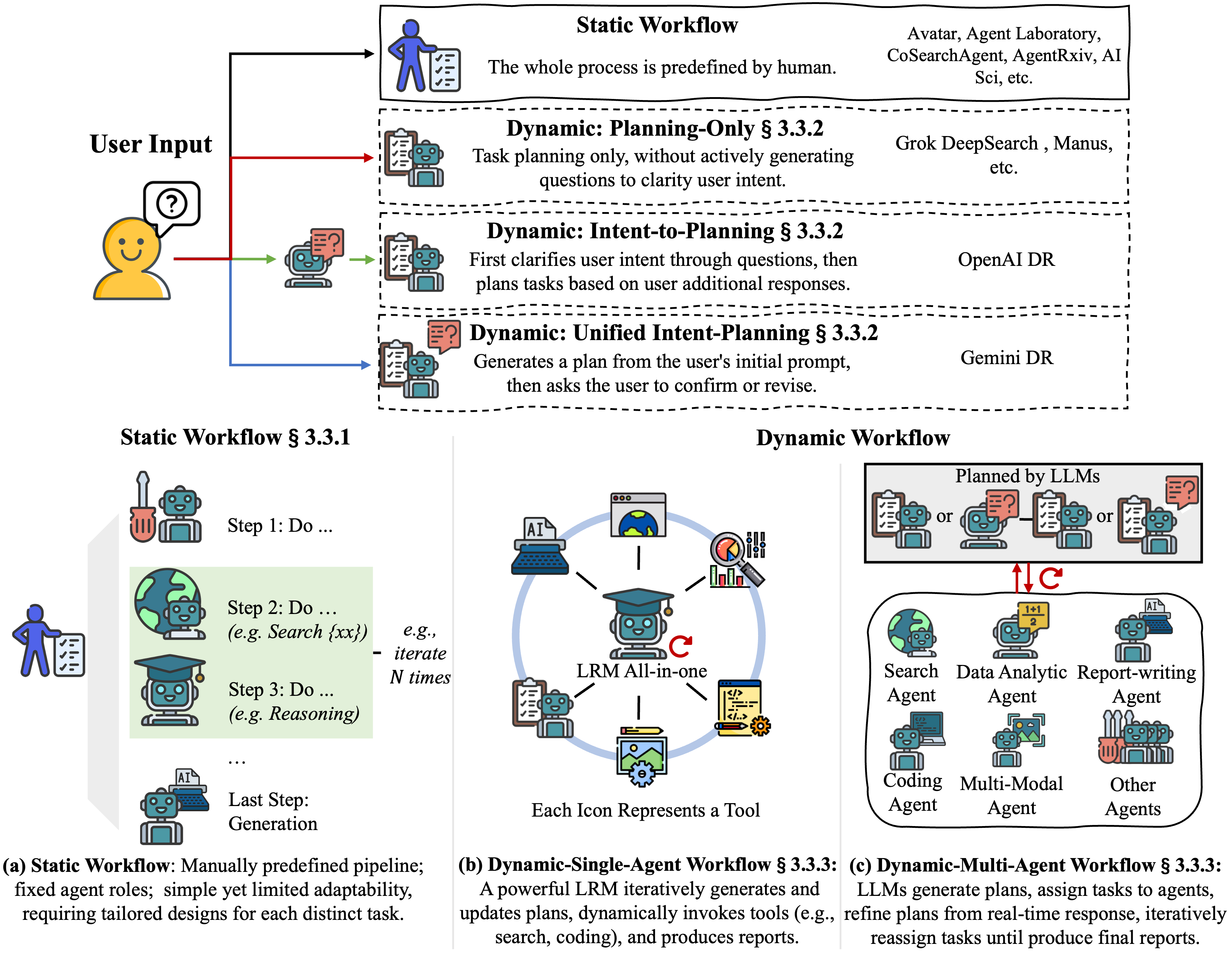}
    \caption{Comparison of DR Workflows: (1) \textbf{Static vs. Dynamic Workflows}: Static workflows rely on predefined task sequences, while dynamic workflows allow LLM-based task planning. (2) \textbf{Planning Strategies}: Three types include: planning-only (direct planning without clarifying user intent), intent-to-planning (clarifying intent before planning), and unified intent-planning (generating a plan and requesting user confirmation). (3) \textbf{Single-Agent vs. Multi-Agent}: Dynamic workflows can be categorised into dynamic-multi-agent systems (tasks distributed across specialised agents) or dynamic-single-agent systems (an LRM autonomously updates and executes tasks). For a continuously updated, per-work taxonomy of DR workflows, see \url{https://github.com/ai-agents-2030/awesome-deep-research-agent}.}
    \label{fig:workflow}
\end{figure}

\subsubsection{Static vs. Dynamic Workflows}
\paragraph{Static Workflows.}
Static workflows rely on manually predefined task pipelines, decomposing research processes into sequential subtasks executed by dedicated agents. These workflows follow explicitly structured procedures, making them particularly suitable for well-defined, structured research scenarios. For instance, AI Scientist \citep{lu2024aiscientistfullyautomated} automates scientific discovery through distinct sequential phases, including ideation, experimentation, and reporting. Similarly, Agent Laboratory \citep{schmidgall2025agent} segments research activities into formalised stages, such as literature review, experimentation, and synthesis of findings. Extending this static paradigm further, AgentRxiv \citep{schmidgall2025agentrxiv} incorporates inter-agent collaboration mechanisms, enabling incremental knowledge reuse through sharing intermediate research outcomes among specialised agents. 

Whist their ease of implementation and structured clarity, static workflows suffer from limited generalisation capabilities, as each distinct task necessitates a specifically tailored pipeline.

\paragraph{Dynamic Workflows.} To overcome the limitations in flexibility and generalizability inherent in static workflows, dynamic workflows support adaptive task planning, allowing agents to dynamically reconfigure task structures based on iterative feedback and evolving contexts. Dynamic architectures leverage advanced mechanisms including automated planning, iterative refinement, and interactive task allocation, enabling tasks to evolve in real-time as new knowledge or external inputs become available. Consequently, dynamic workflows exhibit superior generality and adaptability, making them highly suitable for complex, knowledge-intensive tasks commonly encountered in AI-driven research scenarios.

\subsubsection{Dynamic Workflows: Planning Strategies}
To enhance DR agents' adaptability in response to evolving user requirements and contexts, existing studies propose three distinctive LLM-based planning strategies, each differing in whether and how they interact with the user to clarify intent:
\begin{enumerate}[label=\arabic*),itemsep=0pt,topsep=0pt,parsep=0pt,partopsep=0pt]
    \item The \textbf{Planning-Only} approach directly generates task plans based solely on initial user prompts without actively engaging in further clarification, adopted by the majority of existing DR agents, including Grok~\citep{grokdeepresearch}, H2O~\citep{h2oai} and Manus~\citep{manus2025}.
    \item The \textbf{Intent-to-Planning} strategy actively clarifies user intent prior to planning through targeted questions, subsequently generating tailored task sequences based on clarified user inputs; this method is utilised by OpenAI DR~\citep{openai2025deepresearch}. 
    \item The \textbf{Unified Intent-Planning} approach synthesises these methods by generating a preliminary plan from the initial prompt, together with interactively engaging the user to confirm or revise the proposed plan. Gemini DR~\citep{geminideepresearch} is representative of this strategy, effectively adopts the strength of user-guided refinement.
\end{enumerate}

\subsubsection{Dynamic Workflows: Single-Agent vs. Multi-Agent}

Dynamic workflows of DR agents can be differentiated based on agent architectures into single-agent and multi-agent frameworks, each exhibiting distinct characteristics concerning task specialisation, coordination complexity, and scalability of execution.

\paragraph{Dynamic Single-Agent Systems.}
Dynamic single-agent systems \textbf{integrate planning, tool invocation, and execution within a unified LRM}, streamlining task management into a cohesive cognitive loop. Single-agent architectures autonomously refine task plans and invoke appropriate tools based on evolving contexts, typically without explicit inter-agent coordination. Compared to multi-agent architectures, single-agent systems enable direct end-to-end reinforcement learning (RL) optimisation across the entire workflow, facilitating smoother and more coherent integration of reasoning, planning, and tool invocation. Systems such as Search-o1~\citep{li2025search}, R1-Searcher~\citep{song2025r1}, DeepResearcher~\citep{zheng2025deepresearcherscalingdeepresearch}, WebDancer~\citep{wu2025webdancer}, WebSailor~\citep{li2025websailor}, PANGU Deepdiver~\citep{shi2025pangu}, Agent-R1~\citep{Agent-R1}, ReSearch~\citep{chen2025learning}, Search-R1~\citep{jin2025search}, WebWatcher~\citep{wu2025webwalker}, MiroRL~\citep{2025mirorl}, Memento~\citep{zhou2025agentfly} and Kimi-Researcher~\citep{kimi2025researcher} exemplify this paradigm through iterative cycles of explicit reasoning, action, and reflection, aligning with the ReAct framework~\citep{yao2023reactsynergizingreasoningacting}. However, this streamlined approach places significant demands on the foundation model’s reasoning capabilities, contextual understanding, and autonomous selection and invocation of tools. Additionally, the tightly integrated nature of single-agent systems may limit modular flexibility, complicating independent scaling or optimisation of individual functional components.

\paragraph{Dynamic Multi-Agent Systems.}
Dynamic multi-agent systems leverage multiple specialised agents to collaboratively execute subtasks generated and dynamically allocated through adaptive planning strategies. These systems typically employ hierarchical or centralised planning mechanisms, wherein a coordinator agent continuously assigns and redistributes tasks based on real-time feedback and replanning. Representative frameworks include OpenManus~\citep{openmanus2025} and Manus~\citep{manus2025}, both adopting hierarchical planner-toolcaller architectures. Similarly, OWL~\citep{owl2025} includes a workforce-oriented model, utilising a central manager agent to orchestrate task distribution among specialised execution agents. Furthermore, Alita~\citep{qiu2025alita} incorporates a self-evolution mechanism into DR agents, allowing the agent to online instantiate and configure new MCP servers tailored to specific tasks and environmental conditions. AWorld~\citep{aworld2025} is an open-source framework for building, orchestrating, and training tools using agents and larger multi-agent systems, offering memory and context services and MCP tool integration for scalable evaluation and self-improvement. Webwalker~\citep{wu2025webwalker} mimics human-like web navigation through an explore-critic paradigm. WebThinker~\citep{Li2025webthinker} uses both executed and auxiliary agents to autonomously search, deeply explore web pages, and draft research reports. Such multi-agent configurations effectively handle complex, parallelizable research tasks, thereby enhancing flexibility and scalability in open-ended research scenarios. Nevertheless, a major current challenge of multi-agent systems lies in the inherent complexity of coordinating multiple independent agents, making it difficult to conduct effective end-to-end reinforcement learning optimisation.

\subsubsection{Memory Mechanism for Long-Context Optimisation}
Memory mechanisms empower DR agents to persistently capture, organise, and recall relevant information across multiple retrieval rounds, thereby reducing redundant queries and improving both the efficiency and coherence of DR tasks. During the DR process, agents typically perform extensive multi-round retrieval, generating hundreds of thousands of tokens (or even millions). Although recent advances in LLMs have significantly expanded context window sizes, current limits still constrain tasks involving extremely long contexts. To address these challenges, DR systems have implemented various optimisations for processing extended contexts. Broadly, these optimisations can be categorised into three main strategies: \textbf{(i) Expanding the Context Window Length; (ii) Compressing Intermediate Steps; (iii) Utilising External Structured Storage for Temporary Results}.

\textbf{Extending the Context Window Length.} It is the simple but intuitively effective approach, exemplified by Google's Gemini model~\citep{geminideepresearch}, which supports a context window of up to one million tokens, supplemented by a RAG setup. Despite its straightforwardness, this method often incurs high computational costs and may lead to inefficiencies in resource utilisation during practical deployments. 

\textbf{Compressing Intermediate Step.} An alternative strategy involves compressing or summarising intermediate reasoning steps, significantly reducing the number of tokens processed by the model and thereby improving both efficiency and output quality. Representative frameworks such as The AI Scientist~\citep{lu2024aiscientistfullyautomated} and CycleResearcher~\citep{weng2024cycleresearcher} pass summarised intermediate results between workflow phases. Further, Search-o1 \citep{li2025search} introduced the concept of ``Reason-in-Documents'', utilising LRMs to compress documents, substantially reducing token volume and enhancing model decision-making efficiency. Meanwhile, WebThinker~\citep{Li2025webthinker} uses an auxiliary model to compress the external information. However, a potential drawback of this approach is the loss of detailed information, potentially impacting the precision of subsequent reasoning. 

\textbf{Utilising External Structured Storage.} This is for preserving and retrieving historical information, enabling DR agents to persistently and efficiently store vast amounts of past context beyond the constraints of the context window, improving memory capacity, retrieval speed, and semantic relevance. Popular open-source frameworks such as Manus~\citep{manus2025}, OWL~\citep{owl2025}, Open Manus~\citep{openmanus2025}, and Avatar~\citep{wu2024avataroptimizingllmagents} utilise external file systems to store intermediate outcomes and historical data for subsequent retrieval. Frameworks like AutoAgent~\citep{tang2025autoagentfullyautomatedzerocodeframework} have developed self-managing modules that leverage vector databases to support scalable memory storage and fast similarity-based lookup. Beyond plain text or vector stores, some works propose more semantically structured memory frameworks: for instance, Agentic Reasoning~\citep{wu2025agentic} employ knowledge graphs to capture intermediate reasoning processes and thereby enhance the precision of information reuse, while Agentrxiv~\citep{schmidgall2025agentrxiv} simulates an academic repository akin to arXiv for storing and retrieving relevant outcomes from other agents. Furthermore, Agent-KB~\citep{tang2025agent} and Alita~\citep{qiu2025alita} construct shared knowledge bases and optimised toolsets for agentic problem-solving. Although these structured approaches offer superior semantic retrieval efficiency and accuracy, they typically entail higher development and maintenance costs due to the need for meticulous data structure design and management.

\subsection{Tuning: Beyond Prompting toward Capability Enhancement}

\begin{longtable}{
    >{\arraybackslash}m{2.7cm}
    >{\centering\arraybackslash}m{0.3cm}
    >{\centering\arraybackslash}m{2.0cm}
    >{\centering\arraybackslash}m{2.8cm}
    >{\centering\arraybackslash}m{2.0cm}
    >{\centering\arraybackslash}m{2.2cm}
    >{\centering\arraybackslash}m{1.2cm}
  }

\caption{Comparison of DR Agents with Tuning Methods}
\label{tab:tuning} \\

\multicolumn{7}{l}{
    \blackbox{} = \textbf{Yes},\ 
    \graybox{} = \textbf{Yes but details unknown},\ 
    \whitebox{} = \textbf{Not present}
} \\
\toprule
\textbf{DR Agent} & \textbf{SFT} & \textbf{RL} & \textbf{Base Model} & \textbf{Data} & \textbf{Reward Design} & \textbf{Release} \\
\midrule
\endfirsthead

\multicolumn{7}{l}{\textit{Table \ref{tab:tuning} continued from previous page}} \\
\toprule
\textbf{DR Agent} & \textbf{SFT} & \textbf{RL} & \textbf{Base Model} & \textbf{Data} & \textbf{Reward Design} & \textbf{Release} \\
\midrule
\endhead

\midrule
\multicolumn{7}{r}{\textit{Continued on next page}} \\
\endfoot

\bottomrule
\endlastfoot

\rowcolor{gray!10}
Gemini DR \cite{geminideepresearch} & \graybox & \graybox & Gemini-2.0-Flash & \whitebox & \graybox & \href{https://gemini.google/overview/deep-research/}{Dec-2024} \\
WebWalker & \whitebox & \whitebox & GPT-4o, Qwen-2.5 7--72B & WebWalkerQA & \whitebox & \href{https://arxiv.org/abs/2501.07572}{Jan-2025} \\
\rowcolor{gray!10}
Grok DeepSearch \cite{grokdeepresearch} & \whitebox & \graybox & Grok3 & \whitebox & \graybox & \href{https://x.ai/news/grok-3}{Feb-2025} \\
OpenAI DR \cite{openai2025deepresearch} & \whitebox & \graybox & GPT-o3 & \whitebox & \graybox & \href{https://openai.com/index/introducing-deep-research/}{Feb-2025} \\
\rowcolor{gray!10}
Agentic Reasoning \cite{wu2025agentic} & \blackbox & \whitebox & DeepSeek-R1, Qwen2.5 & GPQA & Rule-Outcome & \href{https://arxiv.org/abs/2502.04644}{Feb-2025} \\
AutoAgent \cite{tang2025autoagentfullyautomatedzerocodeframework} & \whitebox & \blackbox & Claude-Sonnet-3.5 & \whitebox & \whitebox & \href{https://arxiv.org/abs/2502.05957}{Feb-2025} \\
\rowcolor{gray!10}
Towards an AI co-scientist & \whitebox & \whitebox & Gemini 2.0 & \whitebox & \whitebox & \href{https://arxiv.org/abs/2502.18864}{Feb-2025} \\
Agent-R1 \cite{Agent-R1} & \whitebox & PPO \cite{schulman2017proximal}, Reinforce++ \cite{hu2025reinforce}, GRPO \cite{shao2024deepseekmath} & Qwen2.5-1.5B-Inst & HotpotQA & Rule-Outcome & \href{https://github.com/0russwest0/Agent-R1}{Mar-2025} \\
\rowcolor{gray!10}
AutoGLM Romination \cite{zhipu2025autoglm} & \graybox & \graybox & GLM-Z1-Air & \whitebox & \graybox & \href{https://autoglm-research.zhipuai.cn/}{Mar-2025} \\
H2O.ai DR \cite{h2oai} & \blackbox & \graybox & h2ogpt-oasst1-512-12b & \whitebox & \graybox & \href{https://h2o.ai/}{Mar-2025} \\
\rowcolor{gray!10}
Copilot Researcher \cite{microsoft_copilot_researcher} & \graybox & \graybox & o3-mini & \whitebox & \whitebox & \href{https://www.microsoft.com/en-us/microsoft-365/blog/2025/03/25/introducing-researcher-and-analyst-in-microsoft-365-copilot/}{Mar-2025} \\
ReSearch \cite{chen2025learning} & \whitebox & GRPO \cite{shao2024deepseekmath} & Qwen2.5-7B-Inst, Qwen2.5-32B-Inst & 2WikiMultiHopQA & Rule-Outcome & \href{https://arxiv.org/abs/2503.19470}{Mar-2025} \\
\rowcolor{gray!10}
R1-Searcher \cite{song2025r1} & \blackbox & Reinforce++ \cite{hu2025reinforce}, GRPO \cite{shao2024deepseekmath} & Qwen2.5-7B-InSt, LLaMA-3.1-8B-Inst & 2WikiMultiHopQA, HotpotQA & Rule-Outcome & \href{https://arxiv.org/abs/2503.05592}{Mar-2025} \\
Search-R1 \cite{jin2025search} & \blackbox & PPO \cite{schulman2017proximal}, GRPO \cite{shao2024deepseekmath} & Qwen2.5-3B/7B, LLaMA3.2-3B-Inst & NQ, HotpotQA & Rule-Outcome & \href{https://arxiv.org/abs/2503.09516}{Mar-2025} \\
\rowcolor{gray!10}
Nouswise \citep{nouswise2025} & \graybox & \graybox & Nouswise & \whitebox & \graybox & \href{https://nouswise.com/homepage}{Mar-2025} \\
DeepResearcher \cite{zheng2025deepresearcherscalingdeepresearch} & \whitebox & GRPO \cite{shao2024deepseekmath} & Qwen2.5-7B-Inst & NQ, HotpotQA & Rule-Outcome & \href{https://arxiv.org/abs/2504.03160}{Apr-2025} \\
\rowcolor{gray!10}
Genspark Super Agent \cite{genspark} & \whitebox & \graybox & Mixture of Agents & \whitebox & \graybox & \href{https://www.genspark.ai/}{Apr-2025} \\
WebThinker \cite{Li2025webthinker} & \blackbox & Iterative Online DPO \citep{rafailov2023direct} & QwQ-32B & Expert Dataset & Rule-Outcome & \href{https://github.com/RUC-NLPIR/WebThinker}{Apr-2025} \\
\rowcolor{gray!10}
SWIRL \cite{goldie2025synthetic} & \whitebox & Offline-RL & Gemma-2-27B & HotPotQA & \whitebox & \href{https://github.com/RUC-NLPIR/WebThinker}{Apr-2025} \\
SimpleDeepSearcher \cite{SimpleDeepSearcher} & \blackbox & PPO \cite{schulman2017proximal} & Qwen-2.5-7B-In, Qwen-2.5-32B-In, Deepseek-Distilled-Qwen-32B, QwQ-32B & NQ, HotpotQA, 2WikiMultiHopQA, Musique, SimpleQA, MultiHop-RAG & Process-based reward & \href{https://github.com/RUCAIBox/SimpleDeepSearcher}{Apr-2025} \\
\rowcolor{gray!10}
PANGU DEEPDIVER \cite{shi2025pangu} & \blackbox & GRPO \cite{shao2024deepseekmath} & Pangu-7B-Reasoner & WebPuzzle & Rule-Outcome & \href{https://arxiv.org/pdf/2505.24332}{May-2025} \\
Tool-Star \citep{dong2025tool} & \blackbox & GRPO \cite{shao2024deepseekmath} & Qwen-2.5 & NuminaMath, HotpotQA, 2WikiMultiHopQA & Rule-Outcome & \href{https://arxiv.org/abs/2505.16410}{May-2025} \\
\rowcolor{gray!10}
WebDancer \citep{wu2025webdancer} & \blackbox & DAPO \citep{yu2025dapo} & Qwen-2.5-7B/32B, QwQ-32B, DeepSeek-R1, GPT-4o & CRAWLQA, E2HQA & Rule-Outcome & \href{https://arxiv.org/abs/2505.22648}{May-2025} \\
O-agents \citep{zhu2025oagents} & \whitebox & \whitebox & GPT-4o, GPT-4.1, Claude-3.7-Sonnet, DeepSeek-R1, Gemini-2.5 & \whitebox & \whitebox & \href{https://arxiv.org/abs/2506.15741}{Jun-2025} \\
\rowcolor{gray!10}
Kimi-Researcher \citep{kimi2025researcher} & \whitebox & REINFORCE & Kimi k1.5/k2 & \whitebox & Rule-Outcome & \href{https://moonshotai.github.io/Kimi-Researcher/}{Jun-2025} \\
WebSailor \citep{li2025websailor} & \blackbox & DUPO & Qwen-2.5-3B/7B/32B/72B & SailorFog-QA & Rule-Outcome & \href{https://arxiv.org/abs/2507.02592}{Jul-2025} \\
\rowcolor{gray!10}
Agent-KB \citep{tang2025agent} & \whitebox & \whitebox & GPT-4o, GPT-4.1, Claude-3.7-Sonnet, o3-mini, Qwen-3 32B, DeepSeek-R1 & \whitebox & \whitebox & \href{https://arxiv.org/abs/2507.06229}{Jul-2025} \\
WebShaper \citep{tao2025webshaper} & \blackbox & GRPO \cite{shao2024deepseekmath} & Qwen-2.5-3B/7B/32B/72B, QwQ-32B & WebShaper & Rule-Outcome & \href{https://arxiv.org/abs/2507.15061}{Jul-2025} \\
\rowcolor{gray!10}
Cognitive Kernel-Pro \citep{wan2025cognitive} & \blackbox & \whitebox & Claude-3.7-Sonnet, CK-Pro-8B & OpenWebVoyager, Multi-hop URLQA, AgentWebQA, WebWalkerQA, DocBench, TableBench & \whitebox & \href{https://arxiv.org/abs/2508.00414}{Aug-2025} \\
WebWatcher \citep{geng2025webwatcher} & \whitebox & GRPO \cite{shao2024deepseekmath} & Qwen-2.5-VL-32B & BrowseComp-VL, Long-tail VQA, Hard VQA & Rule-Outcome & \href{https://arxiv.org/abs/2508.05748}{Aug-2025} \\
\rowcolor{gray!10}
MiroRL \citep{2025mirorl} & \blackbox & GRPO \cite{shao2024deepseekmath} & Qwen3-14B & MiroRL-GenQA & Rule-Outcome & \href{https://github.com/MiroMindAI/MiroRL}{Aug-2025} \\

\end{longtable}

\paragraph{Parametric Approaches.}  
Prompt-based methods directly leverage the capabilities of pre-trained LLMs, enabling complex functionalities without expensive fine-tuning or additional training. However, it remains challenging to systematically optimise prompt structures and workflows. Moreover, since an agent’s performance is inherently limited by its backbone LLM, increasing the complexity of decision-making processes quickly reaches the model’s performance ceiling. To overcome these limitations, it is essential to incorporate advanced optimisation techniques such as fine-tuning, reinforcement learning (RL) or hybrid training paradigms to further extend the model’s inherent capabilities. Below, we discuss the two main tuning paradigms, supervised fine-tuning (SFT) and RL, and highlight how each extends agent capabilities beyond prompt-only methods.

\subsubsection{SFT-based Optimization}
Prompt-based approaches, while effective for rapid adaptation, are fundamentally constrained by the intrinsic generalisation capacity of backbone LLMs and often exhibit limited robustness in complex task settings. In order to address these limitations, researchers have increasingly explored fine-tuning methodologies aimed at systematically optimising LLMs for critical components of deep research agents. These components include search query formulation, structured report generation, and external tool utilisation. These efforts aim to enhance retrieval quality, mitigate hallucinations, and enable more reliable long-form and evidence-grounded generation.\\\\
An early milestone in this research direction is Open-RAG \citep{islam2024open}, which augments data construction with diverse supervisory signals, including retrieval tokens, relevance tokens, grounding tokens, and utility tokens. Through adversarial training, Open-RAG improves the model's capability to filter irrelevant information, thereby enhancing both retrieval accuracy and the quality of downstream tasks. Building upon this foundation, AUTO-RAG \citep{yu2024auto} enhances the autonomous iterative retrieval capabilities of LLMs. In contrast to earlier multi-hop retrieval approaches that relied on few-shot prompting or hand-crafted templates \citep{jiang2023active,feng2024retrieval,wang2024llms}, AUTO-RAG constructs reasoning-grounded instruction datasets, enabling models to autonomously plan retrieval queries and engage in multi-round interactions with retrievers. The model dynamically refines its retrieval strategy during generation, gathering sufficient evidence before synthesising a final answer. Extending these retrieval-centric innovations, DeepRAG \citep{guan2025deeprag} proposes a binary tree search mechanism that recursively generates sub-queries and constructs multi-turn retrieval trajectories. This mechanism enables the model to judiciously balance between internal parametric knowledge and external retrieval-based rollouts. Consequently, it enhances search efficiency and mitigates redundant external queries.\\\\

In order to further reduce reliance on manually constructed supervised fine-tuning (SFT) datasets, recent work has sought to reduce dependence on manually constructed supervised fine-tuning datasets by developing fine-tuning strategies based on rejection sampling. CoRAG \citep{wang2025chainofretrievalaugmentedgeneration} uses rejection sampling to extract intermediate retrieval chains from standard question answering datasets, allowing for stepwise retrieval augmentation and dynamic reformulation of subqueries as context evolves instead of supervising only final outputs. Li et al. \citep{li2025start} propose a hint-infer mechanism that monitors token patterns during generation and triggers external computational tools, such as Python executors or hint libraries, when specific cues are detected. After an initial supervised fine-tuning phase, the model undergoes a rejection sampling fine-tuning process that teaches it to generate its own prompts and invoke tools autonomously without reliance on hand-curated demonstrations. ATLAS \citep{chen2025atlas} proposes a novel approach for LLM-based agents that trains exclusively on selected critical steps from expert trajectories, significantly improving generalisation performance.\\\\

Although these SFT methods enhance the generalisation of deep research agents by supporting dynamic retrieval planning, structured information synthesis, and integrated tool use, they remain confined to offline, static retrieval pipelines characteristic of retrieval-augmented systems. In contrast, reinforcement learning offers a more adaptive solution for online query generation and tool invocation. By learning from real-time reward signals, reinforcement learning agents acquire the ability to formulate effective search queries and determine the optimal timing for tool calls. This approach addresses the limitations of synthetic demonstration data and distributional shifts, yielding more robust and adaptive performance in open-ended research environments.

\subsubsection{Reinforcement Learning-based Optimisation} RL-based methods optimise DR agents by directly enhancing their adaptive capabilities and generalisation across diverse tasks, surpassing conventional instruction-following or pattern learning approaches. Recent advances have demonstrated that end-to-end RL training significantly strengthens iterative information retrieval, dynamic tool invocation, and integrated reasoning capabilities within DR agents. See comparative analysis in Table~\ref{tab:tuning}.


Early RL-based approaches such as DeepRetrieval~\citep{jiang2025deepretrieval} optimised query generation for improved information retrieval quality, effectively enhancing downstream text generation by producing more relevant search results. Building on query optimisation, ReSearch~\citep{chen2025learning} extended RL to adaptive reasoning over retrieved information. The model dynamically refined search strategies and iteratively updated results based on continuous feedback, significantly improving task-solving accuracy. Subsequently, R1-Searcher~\citep{song2025r1} further optimised retrieval interactions, explicitly training models to refine search strategies through carefully designed reward functions. This allowed better exploitation of external information and improved search result relevance.

Search-R1~\citep{jin2025search} advanced RL-based retrieval by structurally integrating sophisticated search interactions with complex reasoning processes. The method systematically bridged query generation and information reasoning, enabling nuanced responses through refined integration of retrieved content. Finally, this research line culminated in the development of Agent-R1~\citep{Agent-R1}, a comprehensive DR framework integrating RL into end-to-end training of LLM agents. Agent-R1 leveraged diverse tools such as APIs, search engines, and databases, achieving autonomous multi-step task execution and dynamic tool coordination. Through RL-driven optimisation across its entire pipeline, Agent-R1 demonstrated advanced capabilities in adaptive planning, iterative execution, and task refinement. Moreover, WebThinker~\citep{Li2025webthinker} integrates a Web Explorer module for dynamic multi-hop web exploration and employs Iterative Online Direct Preference Optimisation~(DPO)  to seamlessly interleave search, navigation, and report drafting during reasoning, while Pangu DeepDiver~\citep{shi2025pangu} builds on the 7B Pangu model pretrained on Huawei’s Ascend NPUs by introducing Search Intensity Scaling (SIS) through a two-phase SFT and RL curriculum, enabling adaptive adjustment of search depth and frequency in open-web environments.

Table~\ref{tab:tuning} reveals three key RL implementation patterns in DR systems: 1) Industrial systems like Gemini DR~\citep{geminideepresearch} and Grok DeepSearch~\citep{grokdeepresearch} employ proprietary RL implementations with undisclosed details, 2) Academic approaches~\citep{chen2025learning, song2025r1} favor modular RL optimization using GRPO~\citep{shao2024deepseekmath} and Reinforce++~\citep{hu2025reinforce} with transparent reward designs, and 3) Emerging hybrid systems like SimpleDeepSearcher~\citep{SimpleDeepSearcher} combine process-based rewards with multi-task training across 6 QA datasets. The table also highlights the prevalence of Qwen2.5 and LLaMA3 model families as preferred base architectures for RL optimisation.

\textbf{Reward Model and Policy Model.} Most current open-source RL implementations of DR agents, including the methods discussed above, commonly adopt rule-based reward models that explicitly define task-specific objectives such as retrieval relevance, information accuracy, or successful tool invocation. To efficiently perform policy optimisation, recent systems have increasingly utilised Proximal Policy Optimisation (PPO) \citep{schulman2017proximal} and Group Relative Policy Optimisation (GRPO) \citep{shao2024deepseekmath}. In particular, GRPO fundamentally reconfigures the advantage estimation paradigm by replacing traditional value functions with group-relative advantage computation. It expands reward space through intra-group normalisation, and sparse binary rewards are transformed into continuous advantage values spanning wider ranges. This expanded signal space provides richer gradient information for policy updates, as evidenced higher high-reward response density compared to PPO.  In addition, GRPO provides a variance suppression mechanism by constraining advantage estimation within dynamically clustered response groups, such as grouping by reasoning depth or tool usage patterns,  reducing policy gradient variance through local standardisation.  In contrast to PPO, GRPO eliminates separate value networks, removing conflicting optimisation objectives between policy and value functions. Empirical measurements show GRPO reduces gradient direction conflicts from 12 to 3 per training epoch, significantly accelerating convergence.  As a result, GRPO outperforms conventional PPO in wider reward distribution coverage, enhancing exploration capacity and faster KL divergence stabilisation during alignment.

\subsection{Non-parametric Continual Learning }
DR agents depend heavily on LRMs and often utilise complex hierarchical workflows. Parameter-based learning approaches such as SFT and RL encounter significant obstacles in this context, including the need to scale model parameters, manage extensive volumes of structured experience data, and design increasingly intricate training algorithms. In contrast,  non-parametric continual learning approaches offer a scalable alternative: agents refine their capabilities at runtime by optimising external memory, workflows, and tool configurations through continuous interaction with the external environment rather than by updating internal weights. This non-parametric continual learning paradigm enables efficient online adaptation with minimal data and computational overhead, making it well-suited to DR agents with complex architectures.\\\\
Non-parametric continual learning approaches, most notably case-based reasoning (CBR), are currently a mainstream method in LLM-driven agent systems. The CBR-based method enables agents to retrieve, adapt, and reuse structured problem-solving trajectories from an external case bank dynamically. Unlike traditional RAG-based methods, which rely on static databases, CBR facilitates online contextual adaptation and effective task-level generalisation. Such flexibility underscores its potential as a scalable and practical optimisation solution for DR agents with complex architecture.  DS-Agent~\citep{guo2024dsagentautomateddatascience} is a pioneering LLM-driven agent that introduced CBR into automated data science workflows, employing approximate online retrieval from a constructed case bank. Similarly, LAM \citep{guo2025optimizing} applies CBR techniques to functional test generation, combining trajectory-level retrieval with LLM planning in a modular system design. Although DS-Agent itself does not
include a learning phase, Agent K~\citep{grosnit2024largelanguagemodelsorchestrating}advances this paradigm with dynamic external case retrieval and reuse guided by a reward-based memory policy, which exemplifies genuine self-evolution enabling continual adaptation
and optimisation without updating model parameters. Focusing on DR agents, AgentRxiv~\citep{schmidgall2025agentrxiv} further extends this
paradigm by enabling autonomous research agents to collaboratively share and access a centralised repository of prior
research outputs. This framework allows LLM agent laboratories to upload and retrieve reports from a shared preprint
server, simulating an online-updating arXiv-like platform, which can be seen as a comprehensive case bank. Such a
system empowers agents to enhance their capabilities and knowledge through contextual adaptation without modifying
their model parameters. 

Compared to prompt-based methods, which encode fixed demonstrations or task heuristics into static input templates, Non-parametric methods enable dynamic retrieval and adaptation of structured trajectories, thereby facilitating continual task generalisation without manual prompt engineering. Relative to RAG, which typically retrieves unstructured textual content from static corpora, CBR operates at the trajectory level and emphasises reasoning-centred memory organisation. A notable example is the Kaggle Grandmaster Agent \citep{grosnit2024largelanguagemodelsorchestrating}, which demonstrates how LLMs equipped with modular reasoning components and persistent memory can achieve expert-level structured problem solving, aligning closely with the CBR paradigm. These characteristics make CBR particularly well-suited for agents requiring procedural adaptation and context-sensitive optimisation across tasks. Except memory-based method, self-evolution can also arise from dynamic infrastructure adaptation. For example, Alita~\citep{qiu2025alita} monitors task requirements and environmental signals to provision and configure new MCP servers at runtime, seamlessly extending and refining its toolset on demand.\\\\
In summary, these self-evolution paradigms in LLM-driven DR agent systems offer substantial promise for structured reasoning and dynamic retrieval and open new pathways for efficient knowledge reuse and continual learning. Although these methods have not yet achieved widespread attention, they address the high data and computational demands inherent to parameter-based approaches and therefore represent an attractive direction for future research and practical deployment.

\begin{figure}[htbp]
    \centering
    \includegraphics[width=\linewidth]{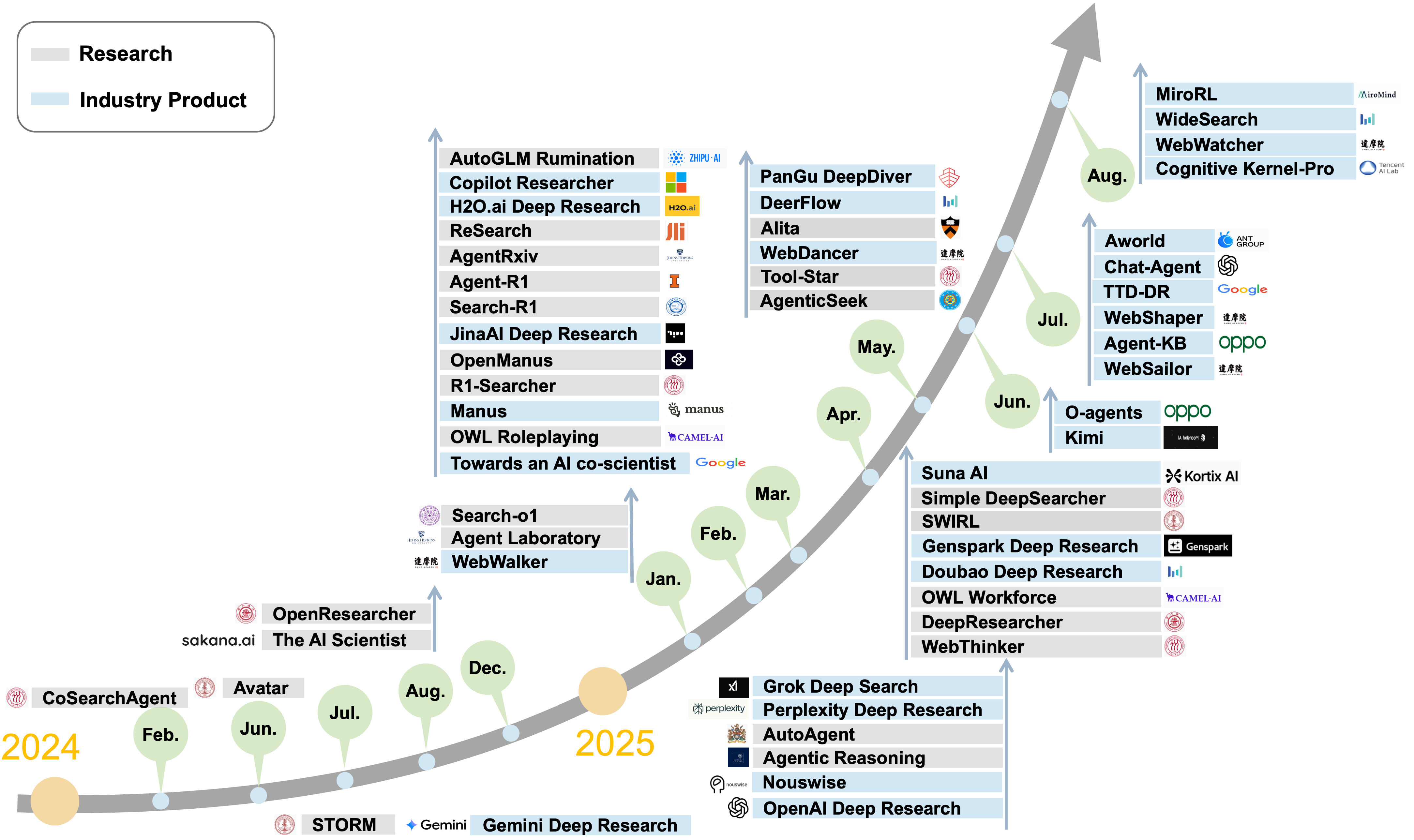}
    \caption{An overview of DR agents' evolution over years.}
    \label{fig:arrow}
\end{figure}

\section{Industrial Applications of Deep Research Agents}

\subsection{Open AI Deep Research}
OpenAI recently introduced its DR capability \citep{openai2025deepresearch}, employing a single-agent architecture centred around a reinforcement learning-based, fine-tuned o3 reasoning model. Upon receiving a research query, the system initiates a concise interactive clarification step to accurately define user intent and research objectives. It then autonomously formulates and executes a sophisticated, multi-step research strategy, encompassing multimodal information retrieval, web browsing, and computational tasks such as data analysis and visualisation through browser tools. Technologically, this solution delivers three significant advancements: (1) \textbf{A dynamically adaptive iterative research workflow}:  Capable of refining its strategy throughout task execution. (2) \textbf{Enhanced context memory and robust multimodal processing capabilities}: Facilitating effective integration of diverse information sources. (3) \textbf{Comprehensive toolchain integration}: Combining web browsing capabilities with built-in programming tools to produce structured, authoritative reports supported by precise citations. 

\subsection{Gemini Deep Research}
Google DeepMind recently introduced Gemini DR \citep{geminideepresearch}, an advanced DR agent based on its multimodal Gemini 2.0 Flash Thinking model. Gemini's reinforcement learning-driven fine-tuning, facilitated by a single-agent architecture, has been shown to enhance planning and adaptive research capabilities, enabling the system to autonomously and expeditiously complete complex tasks. Technologically, this solution delivers four significant advancements: (1) \textbf{Interactive Research Planning}: Upon receiving a research query, Gemini autonomously formulates a multi-step investigation plan for interactive user review and modification. (2) \textbf{Asynchronous Task Management}: Adopts an asynchronous task management architecture to efficiently handle multiple simultaneous tasks. (3) \textbf{Large-scale context windows RAG ensembles}: Enabling effective management and coherent synthesis of multimodal data (eg, text, images)for in-depth professional research analysis. (4) \textbf{High speed adaptive retrieval}: Implements fast, multi-round adaptive web search that significantly outperforms other agents in terms of retrieval speed and amount of information per iteration.

\subsection{Perplexity Deep Research}
Perplexity's recently developed DR agent \citep{perplexitydeepresearch} has demonstrated an advanced capability to decompose complex queries into well-defined subtasks. The system is capable of conducting targeted web searches iteratively, critically evaluating authoritative sources, and synthesising structured, comprehensive reports. Technologically, this solution delivers two significant advancements: (1) \textbf{Iterative Information Retrieval}: Conducts successive rounds of targeted web searches with dynamic adjustments based on interim insights, ensuring comprehensive information coverage and accuracy. (2) \textbf{Dynamic Prompt-Guided Model Selection}: Use a hybrid architecture to autonomously select the optimal combination of specialised models based on the requirements and context of specific tasks, thereby enhancing adaptability and effectiveness in various research scenarios.

\subsection{Grok DeepSearch}
Grok DeepSearch \citep{grokdeepresearch}, developed by xAI, is a computational framework that combines real-time information retrieval with multimodal reasoning to dynamically solve complex and information-rich problems. Technologically, this solution delivers two significant advancements: (1) \textbf{Segment-level module processing pipeline}: Upon receiving a query, Grok3 initiates the credibility assessment module to identify and filter out low-quality information. Subsequently, the system's real-time data acquisition engine gathers multimodal inputs (e.g. text, images, and code) from various sources. Subsequently, employing the sparse attention mechanism, the system undertakes key reasoning subtasks, including data cleaning, cross-source verification, and multimodal integration, in a concurrent manner. Finally, the iterative optimisation process culminates in the generation of structured outputs, encompassing analysis summaries, advanced visualisations (e.g. 3D trajectories), and verifiable citations. (2) \textbf{Dynamic resource allocation}: Capacity for adaptively alternating between lightweight retrieval and intensive analysis modes is noteworthy, and it is further augmented by the incorporation of a secure sandbox environment for computational verification.

\subsection{Microsoft Copilot Researcher and Analyst}
Microsoft recently introduced two innovative reasoning agents within Microsoft 365 Copilot: Researcher and Analyst~\citep{spataro_introducing_2025}. These agents securely and compliantly access users' work data (such as emails, meeting notes, documents, and chats) as well as web information, delivering on-demand expert knowledge.

Researcher is designed to assist users in tackling complex, multi-step research tasks, delivering insights with unprecedented quality and accuracy. It combines OpenAI's advanced research models with Microsoft 365 Copilot's sophisticated orchestration and deep search capabilities. Users can employ Researcher to craft detailed market entry strategies, identify market opportunities for new products by integrating internal and external data, or prepare comprehensive quarterly reports for client reviews. Additionally, Researcher enhances its insights through connectors to third-party data sources such as Salesforce, ServiceNow, and Confluence.

Analyst is built as an advanced data analytics agent that rapidly transforms raw data into valuable insights within minutes. It leverages OpenAI's o3-mini inference model, specifically optimised for advanced analytical tasks in professional environments. Analyst uses a chain-of-thought reasoning approach, solving problems step-by-step, generating high-quality responses that closely mirror human analytical thinking.

\subsection{Qwen Deep Research}
Alibaba Qwen recently launched Qwen Deep Research, an advanced research agent powered by its flagship multimodal model Qwen3-235B-A22B. Through reinforcement learning-optimised task scheduling within a unified agent framework, the system demonstrates enhanced autonomous planning and adaptive execution capabilities, enabling rapid completion of complex research workflows. Key technological advancements include: (1) \textbf{Dynamic Research Blueprinting} with interactive plan refinement. (2) \textbf{Concurrent Task Orchestration} enabling parallel retrieval validation synthesis.

\subsection{Kimi K2 Deep Research}
Moonshot AI’s Kimi K2 \citep{team2025kimi} advances deep research through token-efficient learning, targeted data engineering, a scalable sparse design, and tool-aligned post-training. Concretely, it delivers four contributions: (1) \textbf{Token-efficient pretraining}: stabilised optimisation with selective attention regularisation for reliable large-scale training. (2) \textbf{Data rewriting}: synthetic rephrasing for knowledge and learning-note style transformation for mathematics, yielding a vetted corpus of about 15.5 trillion tokens across web text, code, math, and knowledge. (3) \textbf{Sparse architecture and systems}: mixture-of-experts with multi-head latent attention and an observed sparsity scaling regularity, supported by flexible parallelism and memory-aware execution on H800 clusters. (4) \textbf{Post-training for agents}: supervised tool-use trajectories with sandboxed code execution followed by reinforcement learning that combines verifiable rewards for objective tasks and rubric-based self-critique for subjective tasks. \\\\

In addition to the pioneering DR services previously discussed, major technology corporations such as Microsoft and ByteDance, alongside emerging startups including Jina AI \citep{jina_deepsearch}, H2O \citep{h2oai}, and Zhipu AI \citep{zhipu2025autoglm}, have also introduced their proprietary DR platforms. The advent of these solutions has spurred considerable global interest, reflected by their rapid proliferation, thereby underscoring both the technological attractiveness and substantial market potential of DR applications. Looking forward, continuous advancements in LLM reasoning, retrieval integration techniques, and multimodal generation are expected to enable DR agents to transcend traditional information retrieval and basic tool invocation tasks. Consequently, DR systems are anticipated to tackle increasingly sophisticated reasoning and complex knowledge-construction challenges, ultimately positioning DR as a foundational technological pillar for next-generation intelligent collaborative research platforms.

\section{Benchmarks for DR Agent}

Evaluating DR agents requires benchmarks that capture their full research workflow, including multi‐step information retrieval, cross‐source synthesis, dynamic tool invocation, and structured evidence‐grounded report generation. Existing evaluations fall into two main categories. \textbf{Question‐Answering (QA)} benchmarks range from single‐turn factual queries to complex research‐style problems, assessing agents’ factual knowledge, domain‐specific reasoning, and ability to locate and integrate relevant information. \textbf{Task Execution benchmarks} evaluate broader capabilities such as long‐horizon planning, multimodal understanding, tool usage, and environment interaction by measuring how well agents carry out end‐to‐end research tasks. Although long‐form generation datasets such as Qasper \citep{dasigi2021dataset} and ELI5 \citep{fan2019eli5} provide tests of extended output coherence, their free‐form nature does not align with the structured evidence‐based reporting expected of DR agents. Consequently, there is a pressing need for specialised benchmarks that reflect the multi‐stage, multimodal characteristics of DR workflows and ensure rigorous and relevant assessment of agent performance across all phases of autonomous research.

\begin{table}[htbp]
  {\scriptsize
  \setlength{\tabcolsep}{6.0pt}  
  \rowcolors{4}{gray!10}{white}
  \centering
\caption{Performance of DR agents on major QA benchmarks. The best performance is highlighted in \textbf{bold}, and the second-best is indicated with an \underline{underline}.}
  \label{tab:qa_benchmarks}
  \begin{tabular}{%
    >{\arraybackslash}m{2.7cm}  
    >{\centering\arraybackslash}m{3.4cm}  
    >{\centering\arraybackslash}m{1.1cm}  
    >{\centering\arraybackslash}m{1.1cm}  
    >{\centering\arraybackslash}m{1.1cm}  
    >{\centering\arraybackslash}m{1.1cm}  
    >{\centering\arraybackslash}m{1.1cm}  
    >{\centering\arraybackslash}m{1.4cm}  
  }
    \multicolumn{7}{l}{\small
      \whitebox{} = \textbf{Not present}
    }\\
    \toprule
    {\small \multirow{2}{*}{\textbf{DR Agent}}} & {\small \multirow{2}{*}{\textbf{Base Model}}} & \multicolumn{5}{c}{\small \textbf{QA Benchmarks}} & {\small \multirow{2}{*}{\textbf{Release}}}\\
     &  & {\small \textbf{Hotpot}} & {\small \textbf{2Wiki}} & {\small \textbf{NQ}} & {\small \textbf{TQ}} & {\small \textbf{GPQA}} &  \\
    \midrule
Search-o1 \citep{li2025search} & QwQ-32B-preview & 57.3 & \textbf{71.4} & \underline{49.7} & \underline{74.1} & 57.9 & \href{https://arxiv.org/abs/2501.05366}{Jan-2025} \\
Agentic Reasoning \citep{wu2025agentic} & DeepSeek-R1, Qwen2.5 & \whitebox & \whitebox & \whitebox & \whitebox & 67.0 & \href{https://arxiv.org/abs/2502.04644}{Feb-2025} \\
Grok DeepSearch \citep{grokdeepresearch} & Grok3 & \whitebox & \whitebox & \whitebox & \whitebox & \textbf{84.6} & \href{https://x.ai/news/grok-3}{Feb-2025} \\
AgentRxiv \citep{schmidgall2025agentrxiv} & GPT-4o-mini & \whitebox & \whitebox & \whitebox & \whitebox & 41.0 & \href{https://arxiv.org/abs/2503.18102}{Mar-2025} \\
R1-Searcher \citep{song2025r1} & Qwen2.5-7B-Base & 71.9 & 63.8 & \whitebox & \whitebox & \whitebox & \href{https://arxiv.org/abs/2503.05592}{Mar-2025} \\
ReSearch \citep{chen2025learning}& Qwen2.5-32B-Inst & 67.7 & 50.0 & \whitebox & \whitebox & \whitebox &\href{https://arxiv.org/abs/2503.19470}{Mar-2025} \\
Search-R1 \citep{jin2025search}& Qwen2.5-7B-Inst & 34.5 & 36.9 & 40.9 & 55.2 & \whitebox & \href{https://arxiv.org/abs/2503.09516}{Mar-2025}\\
DeepResearcher \citep{zheng2025deepresearcherscalingdeepresearch} & Qwen2.5-7B-Inst & 64.3 & \underline{66.6} & \textbf{61.9} & \textbf{85.0} & \whitebox & \href{https://arxiv.org/abs/2504.03160}{Apr-2025} \\
WebThinker \citep{Li2025webthinker} & QwQ-32B & \whitebox & \whitebox & \whitebox & \whitebox & \underline{68.7} & \href{https://github.com/RUC-NLPIR/WebThinker}{Apr-2025} \\
SimpleDeepSearch \citep{SimpleDeepSearcher} & QwQ-32B & \textbf{73.5} & \whitebox & \whitebox & \whitebox & \whitebox & \href{https://github.com/RUCAIBox/SimpleDeepSearcher}{Apr-2025} \\
SWIRL \citep{goldie2025synthetic} & Gemma-2-27B & \underline{72.0} & \whitebox & \whitebox & \whitebox & \whitebox & \href{https://arxiv.org/abs/2504.04736}{Apr-2025} \\
Tool-Star \citep{dong2025tool} & Qwen2.5-3B & 51.9 & 40.0 & \whitebox & \whitebox & \whitebox & \href{https://arxiv.org/abs/2505.16410}{May-2025} \\
    \bottomrule
  \end{tabular}}
\end{table}

\begin{table}[htbp]
  {\scriptsize
  \setlength{\tabcolsep}{7.2pt}
  \rowcolors{5}{gray!10}{white}
  \centering
  \caption{Performance of DR agents on GAIA test and validation sets. The best performance is highlighted in \textbf{bold}, and the second-best is indicated with an \underline{underline}.}
  \label{tab:gaia_hle_benchmarks}
  \begin{tabular}{%
    >{\arraybackslash}m{3.3cm}  
    >{\centering\arraybackslash}m{2.8cm}  
    >{\centering\arraybackslash}m{1.3cm}  
    >{\centering\arraybackslash}m{1.3cm}  
    >{\centering\arraybackslash}m{1.3cm}  
    >{\centering\arraybackslash}m{1.3cm}  
    >{\centering\arraybackslash}m{1.6cm}  
  }
      
    \multicolumn{5}{l}{\small
      \whitebox{} = \textbf{Not present}
    }\\
    
    \toprule
    {\small \multirow{2}{*}{\textbf{DR Agent}}} & {\small \multirow{2}{*}{\textbf{Base Model}}} & \multicolumn{4}{c}{\small \textbf{GAIA}} & {\small \multirow{2}{*}{\textbf{Release}}}\\
     &  & {\small \textbf{Level-1}} & {\small \textbf{Level-2}} & {\small \textbf{Level-3}} & {\small \textbf{Ave.}} &  \\
    \midrule
    \textbf{Test set} &  &  &  &  &  & \\
MMAC-Copilot \citep{song2024mmac} & GPT-3.5, GPT-4 & 45.16 & 20.75 &  6.12 & 25.91 & \href{https://arxiv.org/abs/2404.18074}{Mar-2024} \\

H2O.ai DR \citep{h2oai} & Claude3.7-Sonnet & \underline{89.25} & \textbf{79.87} & \textbf{61.22} & \textbf{79.73} & \href{https://h2o.ai/}{Mar-2025} \\

Alita \citep{qiu2025alita} & Claude-Sonnet-4, GPT-4o & \textbf{92.47} & \underline{71.7} & \underline{55.1} & \underline{75.42} & \href{https://arxiv.org/pdf/2505.20286}{May-2025} \\
Agent-KB \citep{tang2025agent} & GPT-4.1, Claude-3.7 & 84.91 & 74.42 & 57.69 & 75.15 & \href{https://arxiv.org/abs/2507.06229}{Jul-2025} \\
O-agents \citep{zhu2025oagents} & Claude-3.7 & 83.02 & 74.42 & 53.85 & 73.93 & \href{https://arxiv.org/abs/2506.15741}{Jun-2025} \\
WebDancer \citep{wu2025webdancer} & QwQ-32B & 61.5 & 50.0 & 25.0 & 51.5 & \href{https://arxiv.org/abs/2505.22648}{May-2025} \\
WebShaper \citep{tao2025webshaper} & Qwen-2.5-72B & 69.2 & 63.4 & 16.6 & 60.1 & \href{https://arxiv.org/abs/2507.15061}{Jul-2025} \\
Deep Researcher with Test-Time Diffusion \cite{han2025deep} & Gemini-2.5-Pro & \whitebox & \whitebox & \whitebox & 69.1 & \href{https://arxiv.org/abs/2507.16075}{Jul-2025} \\
Cognitive Kernel-Pro \citep{wan2025cognitive} & Claude-3-7  & 83.02 & 68.60 & 53.85 & 70.91 & \href{https://arxiv.org/abs/2508.00414}{Aug-2025} \\

\hline
\textbf{Dev set} &  &  &  &  &  & \\
    
AutoAgent \citep{tang2025autoagentfullyautomatedzerocodeframework} & Claude-Sonnet-3.5 & 71.7 & 53.5 & 26.9 & 55.2 & \href{https://arxiv.org/abs/2502.05957}{Feb-2025} \\
OpenAI DR \citep{openai2025deepresearch} & GPT-o3-customized & 78.7 & \textbf{73.2} & 58.0 & 67.4 & \href{https://openai.com/index/introducing-deep-research/}{Feb-2025} \\
Manus \citep{manus2025} & Claude3.5, GPT-4o & \underline{86.5} & 70.1 & 57.7 & \underline{71.4} & \href{https://manus.im/}{Mar-2025} \\
OWL \citep{owl2025} & Claude-3.7-Sonnet & 84.9 & 68.6& 42.3 & 69.7 & \href{https://github.com/camel-ai/owl}{Mar-2025} \\
H2O.ai DR \citep{h2oai} & h2ogpt-oasst1-512-12b & 67.92 & 67.44 & 42.31 & 63.64 & \href{https://h2o.ai/}{Mar-2025} \\
Genspark Super Agent \citep{genspark} & Claude 3 Opus & \textbf{87.8} & \underline{72.7} & \underline{58.8} & \textbf{73.1} & \href{https://www.genspark.ai/}{Apr-2025} \\
WebThinker \citep{Li2025webthinker} & QwQ-32B &53.8 & 44.2 & 16.7 & 44.7 & \href{https://github.com/RUC-NLPIR/WebThinker}{Apr-2025} \\
SimpleDeepSearch \citep{SimpleDeepSearcher} & QwQ-32B & 50.5 & 45.8 & 13.8 & 43.9 & \href{https://github.com/RUCAIBox/SimpleDeepSearcher}{Apr-2025} \\
Alita \citep{qiu2025alita} & Claude-Sonnet-4, GPT-4o & 75.15 & \whitebox & \textbf{87.27} & \whitebox & \href{https://arxiv.org/pdf/2505.20286}{May-2025} \\

    \bottomrule
  \end{tabular}}
\end{table}

\paragraph{QA Benckmarks.}

\begin{table}[!htbp]
{\scriptsize
\setlength{\tabcolsep}{8pt} 
\rowcolors{3}{gray!10}{white}
\centering
\caption{Overview of nine widely used QA benchmark datasets employed in recent DR-agent studies. The first group covers single-hop QA tasks, while the second group focuses on multi-hop and multi-turn reasoning.}
\label{tab:qa_benchmark_diff}
\begin{tabular}{%
  >{\arraybackslash}m{2.8cm}
  >{\centering\arraybackslash}m{0.7cm}
  >{\centering\arraybackslash}m{0.8cm}
  >{\arraybackslash}m{5.2cm}
  >{\arraybackslash}m{1.6cm}
  >{\centering\arraybackslash}m{2.0cm}}
\toprule
{\small \multirow{2}{*}{\textbf{Benchmark}}} &
{\small \multirow{2}{*}{\textbf{Release}}} &
{\small \multirow{2}{*}{\textbf{Size}}} &
{\small \multirow{2}{*}{\textbf{Task \& Context}}} &
{\small \multirow{2}{*}{\textbf{Domain}}} &
{\small \multirow{2}{*}{\textbf{Multi-hop Nums}}} \\
& & & & & \\   
\midrule
TriviaQA \citep{joshi2017triviaqalargescaledistantly} & 2017 & 95 k & Single-hop retrieval (Long web/Wiki docs) & Open & 1 \\
Natural Questions \citep{kwiatkowski2019natural}  & 2019 & 307 k & Document answer extraction (Full Wikipedia article) & Open & 1 \\
PopQA \citep{mallen2023trustlanguagemodelsinvestigating} & 2023 & 14 k & Single-hop parametric recall (None) & Open & 1 \\
TELEQnA \citep{maatouk2023teleqnabenchmarkdatasetassess} & 2023 & 10 k & Domain factual QA (Telecom standards/articles) & Telecom & 1 \\
SimpleQA \citep{wei2024measuringshortformfactualitylarge} & 2024 & 4.3 k & Single-hop factual recall (None / parametric) & Open & 1 \\
\midrule
HotpotQA \citep{yang2018hotpotqadatasetdiverseexplainable} & 2018 & 113 k & Multi-hop reasoning (2 Wikipedia paragraphs) & Open & 2 \\
2WikiMultihopQA \citep{ho2020constructingmultihopqadataset} & 2020 & 192 k & Multi-hop reasoning (Retrieval across Wikipedia) & Open & 2+ \\
Bamboogle \citep{aksitov2023restmeetsreactselfimprovement} & 2023 & 125 & Compositional reasoning (Online search) & Open & 2–3 \\
Humanity’s Last Exam \citep{phan2025humanity} & 2025 & 2.5 k & Expert-level multi-turn (Mixed external sources) & Multi-discipline & 2+ \\
\bottomrule
\end{tabular}}
\end{table}

QA benchmarks span a spectrum of complexity, from simple factual recall to multi-hop reasoning and research-style question answering. At the lower end, datasets such as \textbf{SimpleQA} \citep{wei2024measuringshortformfactualitylarge}, \textbf{TriviaQA} \citep{joshi2017triviaqalargescaledistantly}, and \textbf{PopQA} \citep{mallen2023trustlanguagemodelsinvestigating}focus on parametric or single-hop factual recall, evaluating whether models can retrieve short factual answers from memory or minimal context. \textbf{Natural Questions (NQ)} \citep{kwiatkowski2019natural}  and \textbf{TELEQnA} \citep{maatouk2023teleqnabenchmarkdatasetassess} add complexity by requiring answer extraction from long documents or domain-specific sources. Benchmarks like \textbf{HotpotQA} \citep{yang2018hotpotqadatasetdiverseexplainable}, \textbf{2WikiMultihopQA} \citep{ho2020constructingmultihopqadataset}, and \textbf{Bamboogle} \citep{aksitov2023restmeetsreactselfimprovement} emphasize multi-hop reasoning and supporting evidence selection across documents. At the highest level of difficulty lies \textbf{Humanity's Last Exam (HLE)}~\citep{phan2025humanity}, which targets expert-level, open-domain scientific questions crafted by leading professors in various fields. These questions often require multi-turn retrieval, complex inference, and even multimodal understanding. Additionally, BrowseComp~\citep{browsercomp} is another challenging benchmark proposed by OpenAI to measure the ability of AI agents to locate hard-to-find information. It retains the answer verifiability of the Simple QA benchmark while filtering out those that can be easily solved by LLMs with web search, thus testing agents' information retrieval and synthesis capabilities. Despite recent advancements, leading DR agents still exhibit suboptimal performance on the HLE and BrowserComp benchmark compared to human experts. This highlights these two benchmarks as the most critical and unresolved challenges in the evaluation of DR agents. 

\paragraph{Task Execution Benchmarks.}
Task execution benchmarks evaluate an agent’s integrated capabilities in tool use, environment perception, and information filtering. These can be grouped into two subcategories. The first category comprises general-purpose assistant tasks such as GAIA \citep{mialon2023gaia}, AssistantBench \citep{yoran2024assistantbenchwebagentssolve}, and Magentic-One \citep{fourney2024magenticonegeneralistmultiagentsolving}. These tasks require agents to plan and execute tool-based workflows (for example, searching, browsing, or form filling) within environments that are open-ended and often web-based. Among them, \textbf{GAIA} has emerged as the most important benchmark, offering diverse, realistic tasks that are easily human-solvable but remain highly challenging for current agents. The second subcategory focuses on \textbf{research and code-oriented tasks}, including \textbf{SWE-bench} \citep{jimenez2024swebenchlanguagemodelsresolve}, \textbf{HumanEvalFix} \citep{muennighoff2024octopackinstructiontuningcode}, \textbf{MLGym} \citep{nathani2025mlgymnewframeworkbenchmark}, \textbf{MLE-bench} \citep{chan2025mlebenchevaluatingmachinelearning}, \textbf{MLBench} \citep{tang2024mlbenchevaluatinglargelanguage}, \textbf{MLAgentBench} \citep{huang2024mlagentbenchevaluatinglanguageagents}, and \textbf{ScienceAgentBench} \citep{chen2025scienceagentbenchrigorousassessmentlanguage}, which test agents on completing machine learning pipelines, repairing real-world code, or replicating scientific experiments. These tasks require long-horizon planning, precise tool invocation, and often code generation and validation. Additionally, benchmarks like \textbf{RE-Bench} \citep{wijk2024rebenchevaluatingfrontierai} and \textbf{RESEARCHTOWN} \citep{yu2024researchtownsimulatorhumanresearch} simulate multi-agent research environments, evaluating how well agents collaborate and iterate in multi-role scientific workflows. \\\\
As DR agents continue to integrate more interactive tools, future evaluation may expand into GUI-based manipulation environments. Benchmarks such as \textbf{OSWorld} \citep{xie2024osworldbenchmarkingmultimodalagents}, \textbf{WebArena} \citep{zhou2024webarenarealisticwebenvironment}, and \textbf{SpaBench} \citep{chen2025spabenchcomprehensivebenchmarksmartphone} allow agents to control applications or web interfaces directly, opening new avenues for testing embodied research capabilities in realistic, user-facing scenarios.

\section{Challenge and Future Directions}

Despite the rapid evolution of DR agents and their demonstrated efficacy in automating multi-step information discovery and synthesis, two overarching challenges persist, defining the roadmap for future innovation. First, the breadth and depth of accessible information remain tightly constrained by reliance on static knowledge repositories or conventional search interfaces. Second, the efficiency and robustness of execution workflows and system architectures are limited by linear planning paradigms and monolithic agent designs. Addressing these challenges will be critical to enabling DR agents to function as truly autonomous, adaptable research assistants capable of navigating complex, heterogeneous data landscapes and orchestrating high-throughput, parallelised reasoning processes.

\paragraph{Broaden Information Source.} To meet the information needs of complex tasks, current DR agents adopt static knowledge bases (such as the RAG method) or rely exclusively on search engines and browsers; the former approach is insufficient, while the latter is confined to publicly available web content, thereby significantly constraining their information‐acquisition capabilities. This inherent limitation renders them incapable of retrieving information concealed behind applications, proprietary interfaces or specialised databases. For example, conventional browsing and search techniques cannot penetrate enterprise software, mobile applications, or subscription-only services, such as the Bloomberg Terminal, thereby precluding access to critical, real-time market intelligence. In order to surmount this limitation, it is imperative to integrate a more granular and extensive range of modular tools via MCPs. This approach enables agents to dynamically access specialised tools and resources beyond the scope of standard browsers or search engines. Such resources may include proprietary applications, databases, or APIs, thereby facilitating the retrieval of previously inaccessible data. Consequently, DR agents have the capacity to deliver more precise, adaptive, and context-aware interactions, thereby effectively fulfilling diverse and complex user requirements.

Following the integration of proprietary APIs and databases, the rate-limiting factor in the workflow shifts from data acquisition to webpage interaction efficiency. Conventional human-centred browsers create a further bottleneck for agents. Because they optimise for visual rendering rather than programmatic control, they suffer from sluggish page loads, fragile element locators that shift with every layout change, and aggressive anti-bot defences that often break automated sessions. These shortcomings translate into high latency, unstable scraping and limited parallelism whenever DR agents try to harvest data at scale. To address this bottleneck, researchers have begun to design \textbf{AI-native browsers} such as Browserbase~\citep{browserbase2024}, Browser Use~\citep{browser_use2024}, Dia, Fellou~\citep{fellou-2025}, and the Comet~\citep{comet-perplexity-2025} from Perplexity. expose a stable, structured DOM view that agents can traverse programmatically~\citep{browserbase2024, browser_use2024, comet-perplexity-2025}. \citep{browserbase2024,fellou-2025} supply explicit API hooks for clicking elements and filling forms, which removes the need for brittle coordinate-based actions. \citep{browserbase2024} further executes pages asynchronously in a headless container, reducing load-time variance and avoiding the overhead of a visible interface. \citep{browserbase2024} embeds a vision–language model that tracks dynamic page changes and automatically resolves login gates and anti-bot challenges. \citep{browser_use2024,comet-perplexity-2025} coordinates dozens of tabs in parallel, allowing DR agents to interact with private dashboards, single-page applications, and interactive visualisations at scale. In combination, these capabilities eliminate the delays and fragility that arise when conventional, human-centred browsers sit between the agent and newly unlocked proprietary data sources. 

\paragraph{Fact Checking.}
To further boost factual accuracy, the latest methods add a structured verification loop and self-reflection abilities on top of multi-step retrieval. Concretely, once an agent has drafted a preliminary answer, it does not rush to deliver a verdict. Instead, it proactively launches cross-checks: it looks for independent sources that confirm the same fact and searches for evidence of contradictions. Grok DeepSearch, for example, follows this strategy—it rates the credibility of every source, inspects consistency through as many as seven layers of depth, and verifies each key claim across multiple origins \citep{grokdeepresearch}. This multi-source cross-validation sharply reduces single-source errors and raises answer reliability. At the same time, agents have begun to reflect on their own reasoning. During inference, they inspect and test intermediate results, much like a human researcher’s reflective thinking. Zhipu’s Rumination model~\citep{zhipu2025autoglm}, for instance, pauses after concluding, keeps searching to check whether that conclusion holds, and only then finalises the answer. Such introspection is typically encouraged by adding correctness-oriented rewards in reinforcement learning. If the model detects conflict or uncertainty, it replans its retrieval strategy and, when necessary, backtracks to revise earlier inferences \citep{openai2025deepresearch}. Through this blend of structured verification and self-reflection, research agents now attain an unprecedented level of rigour in fact-checking: they not only supply an answer but also explain why it is trustworthy, dramatically lowering factual errors and hallucinations. In short, modern agents can lay out a search plan, adapt queries as intermediate evidence comes in, and—where needed—rewind prior steps to recover missing information \citep{openai2025deepresearch}.

\paragraph{Asynchronous Parallel Execution.} To address the limitation that most existing DR agents rely exclusively on linear task planning, i.e. the sequential execution of subtasks, we introduce two possible methodologies. These methods overcome the inherent efficiency and robustness constraints of purely linear strategies and enable both the exploitation of parallelism and the implementation of dynamic adjustments during task execution. Firstly,  an asynchronous, parallel architecture leveraging advanced task-modelling structures, such as directed acyclic graphs (DAGs), presents a promising future direction which could enable parallel execution and dynamic prioritisation of subtasks, effectively managing complex interdependencies among tasks and facilitating potentially sophisticated planning capabilities such as replanning. Secondly, a learned scheduling agent, trained via reinforcement learning to allocate subtasks and adjust execution order based on runtime performance signals (e.g. execution latency), could be proposed. By treating scheduling decisions as actions in an RL environment, the agent progressively discovers policies that balance parallelism, resource utilisation, and task criticality, yielding more robust and efficient end-to-end research workflows.

\paragraph{Tool-Integrated Reasoning.}
A fundamental challenge in developing effective DR agents lies in the implementation of Tool-Integrated Reasoning (TIR), a paradigm that extends beyond simple tool usage to encompass complex, multi-step reasoning with dynamic tool integration. TIR requires agents to not only invoke appropriate tools in logical sequence but also to adaptively adjust their reasoning pathways based on intermediate results. Traditional supervised fine-tuning approaches have demonstrated limited generalisation capabilities in tool-based reasoning tasks, often leading to over-reasoning or inappropriate tool selection. Recent research by \citep{qiancheng2025toolrl} has shown that reinforcement learning frameworks with carefully designed reward structures can significantly enhance models' tool reasoning abilities. By incorporating fine-grained rewards that evaluate not only final answer correctness but also tool selection appropriateness, parameter specification accuracy, and reasoning efficiency, TIR-optimised agents have demonstrated performance improvements of 15-17\% across multiple benchmarks. Furthermore, these agents exhibit superior generalisation to unseen tools and tasks, more rational invocation patterns, and better balance between tool utilisation and self-knowledge. Implementing TIR effectively within DR agents represents a critical step toward achieving truly autonomous research assistants capable of navigating complex information landscapes with minimal human intervention.
 
\paragraph{Benchmark Misalignment.}
Most public DR evaluations remain anchored in traditional QA suites whose items are harvested chiefly from static corpora such as Wikipedia. Since a considerable amount of this content is now embedded in backbone model parameters, current competitive agents can often answer directly from memory, bypassing any research procedure and thus inflating their performance. To probe genuine capabilities of retrieval, reasoning and tool usage, the field of DR urgently needs open-web, time-sensitive benchmarks. From this perspective, BrowseComp~\citep{browsercomp} constitutes a meaningful step forward by filtering out questions solvable with parametric knowledge and forcing agents to locate hard-to-find information online. Besides, a complementary direction is a continually refreshed leaderboard that updates problems from the latest web environment and events, deterring benchmark hacking through parametric memorisation.\\\\
Beyond parametric knowledge hacking of QA benchmark, the metrics of the most existing DR research still collapse open-ended research workflows into narrowly scoped QA prompts or rudimentary GUI-control tasks, overlooking the paradigm’s defining outcome, a structured, multi-modal research report that weaves together textual narrative, tables, figures, and citations. Since the metrics of these benchmarks centre almost exclusively on information retrieval and extraction and tool invocation, they under-assess higher-level competencies such as evidence aggregation across heterogeneous sources, cross-modal synthesis, and discourse-level organisation. Thus, a key research priority is the development of comprehensive benchmarks that evaluate DR agents’ capacity for end-to-end report generation, encompassing long-form narrative, integrated tables and figures, and multimodal coherence, thereby assessing factual accuracy, discourse structure, and cross-modal alignment within a single task.

\paragraph{Parametric Optimisation of Multi-Agent Architectures.}
End-to-end RL has been demonstrated by OpenAI~\citep{openai2025deepresearch, Agent-R1} to significantly enhance the reasoning capabilities of backbone models for DR tasks, a result successfully replicated by several open-source initiatives. However, current implementations predominantly utilise single-agent architectures, requiring the backbone model to simultaneously manage planning, tool invocation, and report generation. This multitasking places excessive computational and cognitive demands on backbone models, thereby reducing their efficiency and robustness. Distributing workloads across multiple specialised agents has shown promising improvements in system performance~\citep{wang2024mobile}, yet achieving effective end-to-end training and efficient coordination among multiple agents remains a critical open challenge.

To optimize multi-agent architectures for DR tasks, we propose two promising future directions: (i) adopting\textbf{ hierarchical reinforcement learning} (HRL), which introduces layered internal reward mechanisms that facilitate efficient feedback propagation and foster cooperative learning among agents; or implementing a post-training optimization pipeline consisting of multiple refinement stages specifically tailored for DR tasks, which could iteratively enhance inter-agent interactions and thus improve overall system stability and adaptability; and (ii) employing an \textbf{RL-based dedicated scheduling agent designed to dynamically allocate subtasks and adjust execution order based on real-time performance metrics}. By modelling scheduling decisions as actions within an RL framework, this method progressively learns adaptive policies that optimally balance parallel execution, resource utilisation, and task prioritisation, enhancing both the robustness and efficiency of end-to-end research workflows.

\paragraph{Self-Evolving Language Model Agents.} Although initial attempts at self-evolution methods for DR agents have emerged, exemplified by simulated collaborative platforms such as AgentRxiv~\citep{schmidgall2025agentrxiv} that facilitate online sharing and reuse of structured research experiences, the paradigm remains underdeveloped and narrowly focused on only the case-based reasoning paradigm. Similarly, CycleResearcher~\citep{weng2024cycleresearcher} enables the entire research process simulation (research-evaluation-refine) through iterative preference learning with a robust verifier~\citep{zhu2025deepreview}, representing a significant step toward fully automated scientific inquiry and sharing the similar self-evolution concept with AlphaEvolve~\citep{novikov2025alphaevolve}.

To fully realise the potential of self-evolution in DR agents, future research should expand the self-evolution method along two complementary directions. (i) \textbf{Comprehensive case-based reasoning framework}. Case-based reasoning approaches~\citep{aamodt1994case} leverage hierarchical experience traces, including planning trajectories and structured tool invocation logs, and employ advanced retrieval and selection mechanisms to enable fine-grained, context-specific adaptation. (ii) \textbf{Autonomous workflow evolution} promises enhanced efficiency and flexibility. By representing agent workflows as mutable structures such as trees or graphs, researchers can apply evolutionary algorithms or adaptive graph optimisation to explore, modify and refine execution plans dynamically. Pursuing both directions in tandem will strengthen the robustness of frameworks and reduce the reliance on data and computation resources.

\section{Conclusion}
LLM-driven Deep Research Agents represent an emerging paradigm for automated research support, integrating advanced techniques such as iterative information retrieval, long-form content generation, autonomous planning, and sophisticated tool utilisation. In this survey, we systematically reviewed recent advancements in DR agents, categorising existing methodologies into prompt-based, fine-tuning-based, and reinforcement learning-based approaches from the perspectives of information retrieval and report generation. Non-parametric methods utilise LLMs and carefully designed prompts to achieve efficient and cost-effective deployment, making them suitable for rapid prototyping. In contrast, fine-tuning and reinforcement learning approaches explicitly optimise model parameters, significantly enhancing the agents' reasoning and decision-making capabilities. We also examined prominent DR agent systems developed by industry leaders and discussed their technical implementations, strengths, and limitations.

\section*{Limitation}
Despite notable progress, key challenges remain, including limited generalisation across diverse tasks, inflexible task workflows, difficulty in integrating granular external tools, and substantial computational complexity associated with advanced planning and optimisation. Future research directions thus emphasise broader and more flexible tool integration through modular capability providers (e.g., Operator-based architectures), development of asynchronous and parallel planning frameworks (e.g., Directed Acyclic Graph-based approaches), and sophisticated end-to-end optimisation methods for multi-agent architectures, such as hierarchical reinforcement learning or multi-stage fine-tuning pipelines. With continued advancements in LLM technologies, DR agents have significant potential to transform complex research workflows, enhance human productivity, and drive innovation across academic and industrial domains.

\bibliographystyle{plain}  
\bibliography{main}

\end{document}